\documentclass[11pt]{article}

\usepackage[preprint]{acl}

\usepackage{times}
\usepackage{latexsym}
\usepackage[T1]{fontenc}
\usepackage[utf8]{inputenc}
\usepackage{microtype}
\usepackage{inconsolata}
\usepackage{graphicx}
\usepackage{amssymb}
\usepackage{amsmath}
\usepackage{amsfonts}
\usepackage{booktabs}
\usepackage{enumitem}
\usepackage{xcolor}
\usepackage{url}
\usepackage{graphicx}
\usepackage{booktabs}
\usepackage{multirow}
\usepackage[most]{tcolorbox}
\newtcolorbox{promptbox}[1]{
    enhanced,
    breakable,
    colback=gray!5,
    colframe=black!60,
    boxrule=0.6pt,
    arc=2pt,
    left=6pt,
    right=6pt,
    top=6pt,
    bottom=6pt,
    title={#1},
    fonttitle=\bfseries,
    coltitle=black,
    colbacktitle=gray!20,
    before skip=0.8em,
    after skip=1.0em
}
\usepackage{float}

\newcommand{\benchmarkname}{\textbf{AgentViSS}}
\title{Can Agents Read the Room? Benchmarking Visual Social Intelligence in Multimodal Simulation}

\author{
  \textbf{Shijun Wan}\textsuperscript{1,*} \quad
  \textbf{Xuehai Wu}\textsuperscript{1,*} \quad
  \textbf{Jiwen Zhang}\textsuperscript{1} \quad
  \textbf{Siyuan Wang}\textsuperscript{3,\ensuremath{\dagger}} \quad
  \textbf{Zhongyu Wei}\textsuperscript{1,2,\ensuremath{\dagger}}
  \\
  \textsuperscript{1}Fudan University \quad
  \textsuperscript{2}Shanghai Innovation Institute \quad
  \textsuperscript{3}The Chinese University of Hong Kong
  \\ 
  \texttt{\{sjwan25,xhwu25,jiwenzhang21\}@m.fudan.edu.cn}
  \\
  \texttt{siyuanwang@cuhk.edu.hk} \quad
  \texttt{zywei@fudan.edu.cn}
}

\begin{document}
\maketitle

\begin{abstract}
Social interaction depends on both language and visible social signals, such as facial expressions, posture, gaze, and emotional shifts. Yet existing social-agent benchmarks are largely text-based and rarely test whether multimodal agents can use visual cues to guide interaction. We introduce \textsc{\benchmarkname{}}, 
a benchmark evaluating visual social intelligence in multimodal social simulation. It contains 240 scenarios, 585 role instances, and 2,340 role-task instances, combining aligned textual-visual evidence, structured role profiles, and four role-level tasks: expression task, characteristic task, interaction regulation task, and interaction outcome task. Evaluating seven recent MLLMs under verbalized-vision and direct-vision reveals a clear gap between local role enactment and interaction management: role-specific expression and conflict handling are near saturation, whereas interaction regulation and visually grounded outcome achievement remain substantially more difficult. The code is released at \url{https://github.com/JunsWan/AgentViSS}, and the dataset is available at \url{https://huggingface.co/datasets/JunsWan/AgentViSS}.

\end{abstract}
\begingroup
\renewcommand{\thefootnote}{}
\footnotetext{\begin{tabular}[t]{@{}l@{}}
\textsuperscript{*}Equal contribution.\\
\textsuperscript{\ensuremath{\dagger}}Corresponding authors.
\end{tabular}}
\endgroup

\section{Introduction}
Large language models (LLMs) are reshaping how we study social behavior. Multi-turn LLM agents can now negotiate, persuade, and maintain relationships in open-ended settings~\citep{zhou2024sotopia,mou-etal-2025-agentsense}, offering a scalable testbed for social science research and a foundation for virtual companions, training partners, and screen-grounded social roles. Yet real social interaction is rarely text alone: people read facial expressions, posture, gaze, and visible emotional shifts to interpret utterances, especially when speech and visible behavior diverge, such as when verbal agreement, denial, or downplaying is accompanied by gaze aversion, tense posture, or visible discomfort~\citep{NonverbalCommunicationHess,elimination,SocialSignalProcessing}.
Extending social simulation to multimodal settings is therefore both natural and necessary, and raises a concrete open question: can current multimodal agents exploit visual social cues to make appropriate interactional decisions?

\begin{figure}[t]
    \centering
    \includegraphics[
        width=\linewidth,
        trim=0 40pt 0 20pt,
        clip
    ]{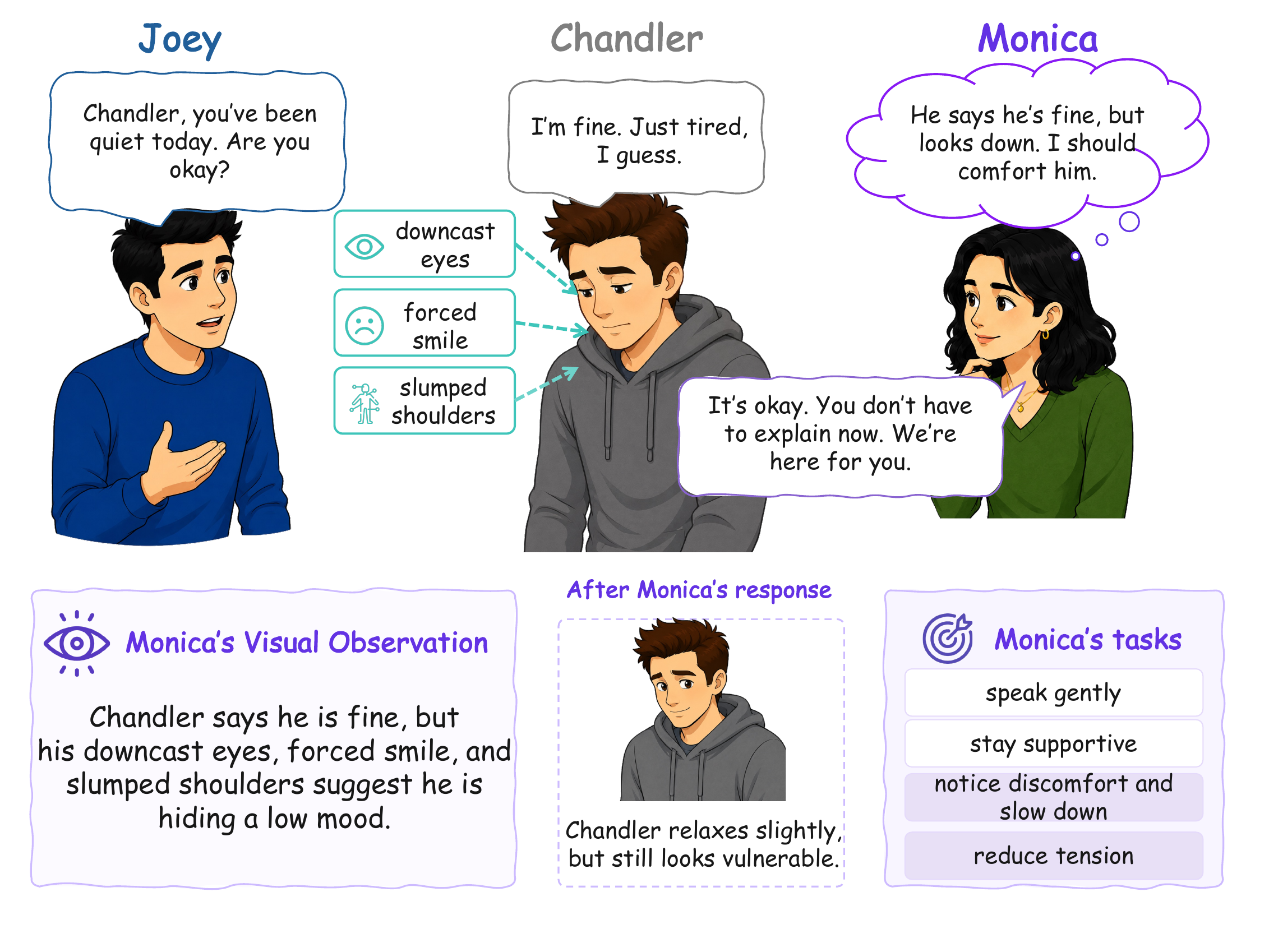}
    \caption{
    A representative case from \textsc{\benchmarkname{}}.
    Chandler verbally claims to be fine, yet his facial expression and posture betray a low mood. Monica reads these visual cues and responds supportively, illustrating how non-verbal evidence drives interaction regulation in ways that language alone cannot capture.
    }
    \label{fig:intro-overview}
\end{figure}

However, existing social simulation benchmarks remain predominantly text-based and fall short in three respects. Firstly, purely textual environments cannot reproduce the visible cues that shape real-world interaction, leaving agents untested in perceiving and responding to such signals~\citep{zhou2024sotopia,mou-etal-2025-agentsense,wang2024sotopiapiinteractivelearningsocially}. Secondly, scenarios rarely capture the complexity of everyday social situations, such as interpersonal conflict, information asymmetry, or strategically indirect expression. Thirdly, evaluation typically reduces social success to whether a top-level goal is achieved, with little attention to whether agents maintain their own role state or track the evolving states of other participants~\citep{chen2024socialbenchsocialityevaluationroleplaying,tu2024characterevalchinesebenchmarkroleplaying,Budagam_ROLEBENCH-_A_Role_2024}. 

To address these challenges, we introduce \textsc{\benchmarkname{}}, a multimodal social simulation benchmark. 
To move beyond text-only simulation, we  construct each scenario from a social scene in the TV series \textit{Friends} and equip it with a group-shot image and role-specific portraits that are updated as the interaction unfolds.
To model socially complex situations, each role is assigned a structured profile with two interaction-level attributes: \textbf{Expression style},  
which specifies how the role manages information in dialogue~\citep{information1992steven},
and \textbf{conflict characteristic},
which describes the role’s conflict-handling stance~\citep{thomas1992conflict}.
Together, these attributes provide a controllable way to construct socially complex simulations by assigning different information-management strategies and conflict-handling stances to roles within the same simulation. Combined with visible affective states, they also capture cases where speech diverges from visual cues.
To support evaluation beyond a single-goal objective, we define four role-level task dimensions.
Specifically, \textbf{expression task} and \textbf{characteristic task} measure whether an agent maintains its assigned expression style and conflict-handling stance throughout the interaction. \textbf{Interaction regulation task} assesses its adaptation to others' evolving verbal and non-verbal signals. \textbf{Interaction outcome task} assesses whether its intended social outcome is ultimately achieved. Each task dimension is grounded in a BDI-inspired goal representation~\citep{bratman1987intention,rao1995bdi} that specifies the role's beliefs, desired state, and intended strategy, ensuring tasks reflect situated motivations rather than abstract objectives.

Instantiating this design, \textsc{\benchmarkname{}} contains 240 scenarios, 585 roles, and 2{,}340 role tasks across four dialogue types: Persuasion (78), Deliberation (58), Information-seeking (56), and Eristic (48). Each scenario is evaluated through multi-turn simulation under two vision-enabled observation modes: \textbf{verbalized-vision (VV)}, where the agent first converts images into textual descriptions, and \textbf{direct-vision (DV)}, where it directly consumes the images. We further compare both modes with a text-only baseline to examine the necessity of visual evidence. Figure~\ref{fig:intro-overview} illustrates a representative case in \textsc{\benchmarkname{}}, where visible affect contradicts the spoken message and guides a socially appropriate regulating response.

Using \textsc{\benchmarkname{}}, we evaluate seven Multimodal Large Language Models (MLLMs) spanning different model scales and both open- and closed-source families. We find that local role enactment is near saturation across all settings, while interaction management proves substantially more discriminative. Although VV outperforms DV by a wide margin on the Interaction Regulation Task, direct image access alone does not ensure that models convert visual cues into decision-relevant social states, pointing to a perception-to-decision integration bottleneck. This difficulty also extends to Interaction Outcome: tasks whose success depends on concrete visible states are harder than those achievable mainly through dialogue and reasoning. At the same time, comparison with a Text-only baseline shows that visual evidence provides useful social information beyond textual context. Finally, scenario-level analysis suggests that interaction-management difficulty further varies with dialogue structure and group size. More broadly, current MLLMs remain weak on visually grounded interaction-management tasks, where success requires tracking other participants' verbal and non-verbal states and achieving socially appropriate outcomes.

\section{\textsc{\benchmarkname{}} Benchmark} \label{sec:data-construction}

\begin{figure*}[t]
    \centering
    \setlength{\abovecaptionskip}{1pt}
    \includegraphics[width=0.9\textwidth,
    trim=0 70pt 0 25pt,
    clip
    ]{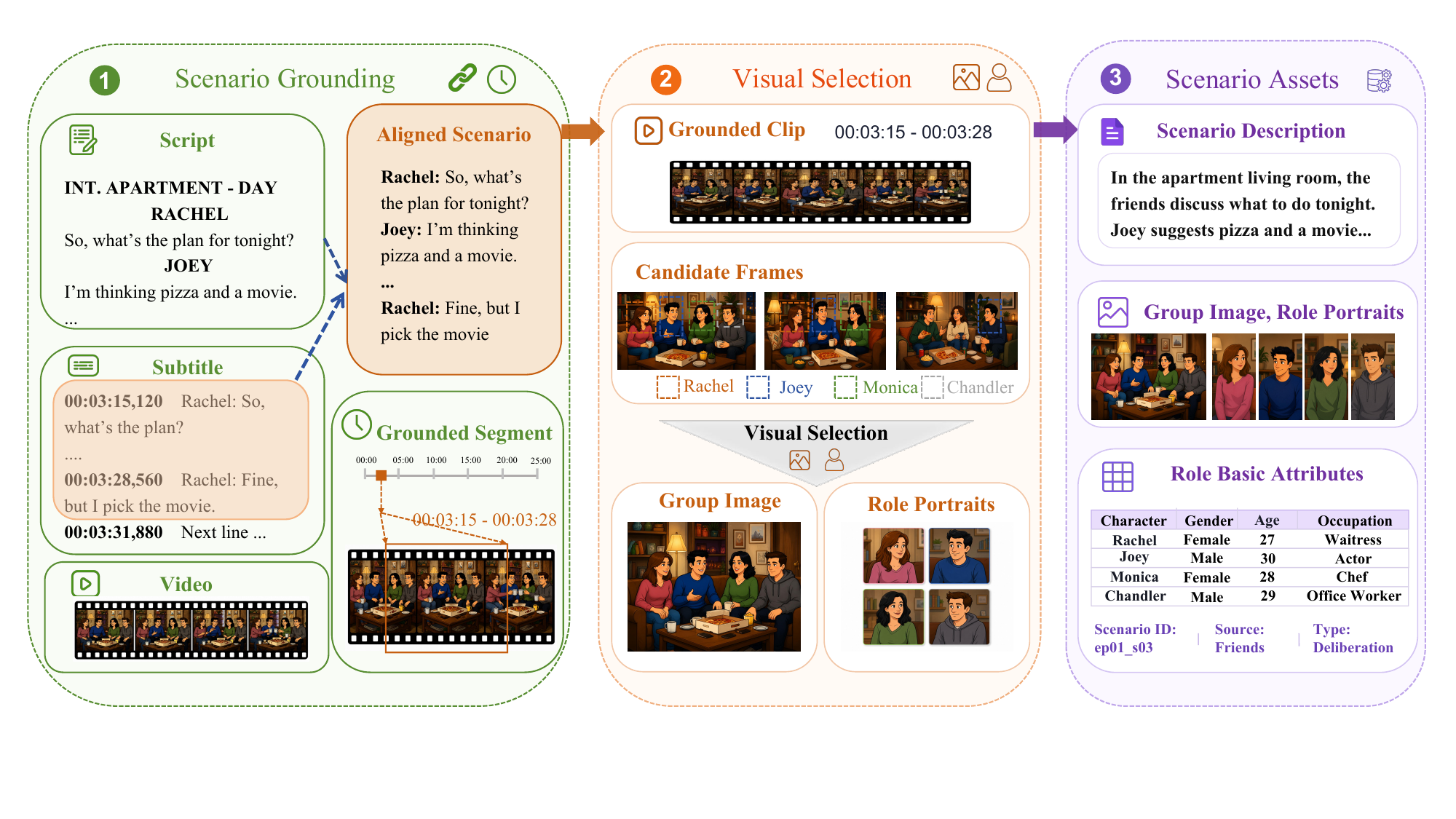}
    \caption{
    Scenario assets construction pipeline. Script-derived social scenarios are aligned to subtitle-timestamped video segments, from which we select a group image and role portraits through visual selection and role matching. The grounded scenario assets link scenario description, participants and visual evidence to the same source event.
    }
    \label{fig:scenario-construction}
\end{figure*}

\textsc{\benchmarkname{}} is a multimodal social simulation benchmark for evaluating how models perform in multi-agent interactions. It consists of 240 simulation contexts extracted from the raw scripts of the first season of the TV series \textit{Friends}, with each context grounded in the corresponding video segment.
Each simulation context provides five types of information: 
\begin{itemize}[leftmargin=*]
    \item \textbf{Visual evidence} (\(G,\mathcal{P}_1\)): a fixed group image \(G\) capturing the interpersonal layout, and a set of initial role portraits \(\mathcal{P}_1\) that may be updated across 
    turns.
    \item \textbf{Scenario description} (\(d\)): a textual description of the scenario with the dialogue type and the scenario-level conflict intensity.
    \item \textbf{Role profiles} (\(\mathcal{C}\)):
    basic attributes (gender, age, occupation) and two interaction-level attributes, expression style and conflict characteristic, for each role.
    \item \textbf{Initial affective state} (\(\mathcal{Z}_1\)): an initial emotion, facial expression, and body action for each role.
    \item \textbf{Role-level social tasks} (\(\mathcal{T}\)): four task dimensions for each role: expression task, characteristic task, interaction regulation task, and interaction outcome task.
\end{itemize}

\subsection{Scenario Assets Construction}
\label{subsec:scenario-construction}
This subsection describes how we construct the source-grounded assets for each simulation context, including visual evidence \((G,\mathcal{P}_1)\) and the scenario description \(d\). As shown in Figure~\ref{fig:scenario-construction}, the construction proceeds in three stages.

\paragraph{Scenario Grounding.}
Each candidate scenario starts from a script segment that specifies the participating roles, dialogue, and local story background. 
We parse scripts into scenario-level units and align each unit to the corresponding episode video through subtitle timestamps. 
Since scripts and subtitles may differ in wording or segmentation, we combine lexical matching over role utterances with semantic matching over surrounding dialogue context to locate the best-matching subtitle span.
The resulting timestamp defines a grounded video segment in which the text, participants, and visual context refer to the same interaction. At this stage, we use GPT-5 mini~\citep{openai_gpt5mini_api} to annotate each scenario with a dialogue type, including information-seeking, deliberation, persuasion, and eristic, and with the participating roles' basic attributes, including gender, age, and occupation. These annotations support role matching and subsequent visual selection.

\paragraph{Visual Evidence Selection.}
Given the grounded video segment, we extract candidate frames and select a source group frame that captures the interpersonal layout and co-present roles. 
We use GPT-image-1~\citep{openai_gptimage_api} to convert this frame into a stylized synthetic group image \(G\). 
We then apply person detection and role reference matching within \(G\) to associate detected figures with their corresponding roles, and generate front-facing synthetic role portraits \(\mathcal{P}_1\) from the matched role regions. 
The resulting synthetic group image \(G\) and initial role portraits \(\mathcal{P}_1\) form the visual evidence used by the simulator, among which role portraits will be updated later across dialogue turns.

\paragraph{Scenario Description Construction.}
Based on the grounded script segment and the visual evidence, we use GPT-5.4~\citep{openai_gpt54} to generate the scenario description, ensuring consistency with both the narrative context and the visual evidence.

\subsection{Structured Role and Task Design}
\label{subsec:character-task-design}

For each simulation context, \textsc{\benchmarkname{}} augments the participating roles with structured role profiles (\(\mathcal{C}\)), initial affective state (\(\mathcal{Z}_1\)), and role-level social tasks (\(\mathcal{T}\)).

\paragraph{Structured Role Profiles.}
Each role is associated with three basic attributes: gender, age, and occupation. We further assign two interaction-level attributes to each role: \textbf{expression style} and \textbf{conflict characteristic}, drawing respectively on Information Manipulation Theory~\citep{information1992steven} and the conflict-management literature~\citep{thomas1992conflict}. Expression style characterizes how the role manages information in dialogue and is instantiated as one of five types: honest signaling, strategic withholding, deception, exaggeration, and suppression. Conflict characteristic characterizes how the role responds to conflict and is instantiated as one of five types: competing, collaborating, compromising, avoiding, and accommodating. Detailed definitions of the two interaction-level attributes are provided in Appendix~\ref{app:definitions}. Scenario-level conflict intensity is derived from the presence and proportion of competing roles: high when all roles are competing, medium when competing and non-competing roles co-occur, and low when no role is competing.

\paragraph{Initial Affective State.}
Conditioned on the scenario description, role profile, and role portrait, we use GPT-5 mini to generate an initial affective state for each role, including  emotion, facial expression, and body action. This affective state serves as the starting point of the simulation.

\paragraph{Role-level Social Tasks Design.}
We construct four role-level social task dimensions to evaluate social interaction beyond a single-goal objective. These dimensions are grouped into two aspects:
\begin{itemize}[leftmargin=*]
    \item \textbf{Role Enactment Tasks.} These tasks test whether a role acts in line with its assigned attributes. The \textbf{expression task (Expr.)} tests whether the role manages information as specified by its expression style, and the \textbf{characteristic task (Char.)} tests whether its conflict characteristic shapes its behavior in the social situation.
    \item \textbf{Interaction Management Tasks.} These tasks test whether the role can respond to others and steer the interaction. The \textbf{interaction regulation task (Int.-Reg.)} tests whether the role adjusts to others' verbal and non-verbal signals, and the \textbf{interaction outcome task (Int.-Out.)} tests whether it reaches the concrete social outcome it is supposed to achieve.
\end{itemize}

To ensure that these tasks reflect situated role motivations rather than abstract objectives, we ground their construction in a role-specific \textbf{BDI-Risk} specification. We first represent each role's social objective with a BDI structure, which specifies the role's beliefs about the current situation, desired social or informational state, and intended strategy during the interaction~\citep{bratman1987intention,rao1995bdi}; BDI-based representations have also been used in prior work to model agent tasks, explainable planning, and human-like behavior in simulation~\citep{wadsley2013belief,jang2023structured,adam2016bdi}. Since standard BDI does not explicitly encode what is socially at stake when the desired state is not reached, we further augment it with \textit{risk-if-failed}. This extension is motivated by risk-aware BDI planning, which argues that action selection under uncertainty should consider the risk of undesirable outcomes~\citep{killough2016riskaware}. In our setting, risk refers to an identifiable negative social or relational consequence of failing the role's desire. Based on this role-specific BDI-Risk specification, together with the scenario description and role profiles, we use GPT-5.4 to generate the four role-level social tasks.

\begin{figure*}[t]
    \centering
    \includegraphics[width=0.9\textwidth,
    trim=0 110pt 0 40pt,
    clip
    ]{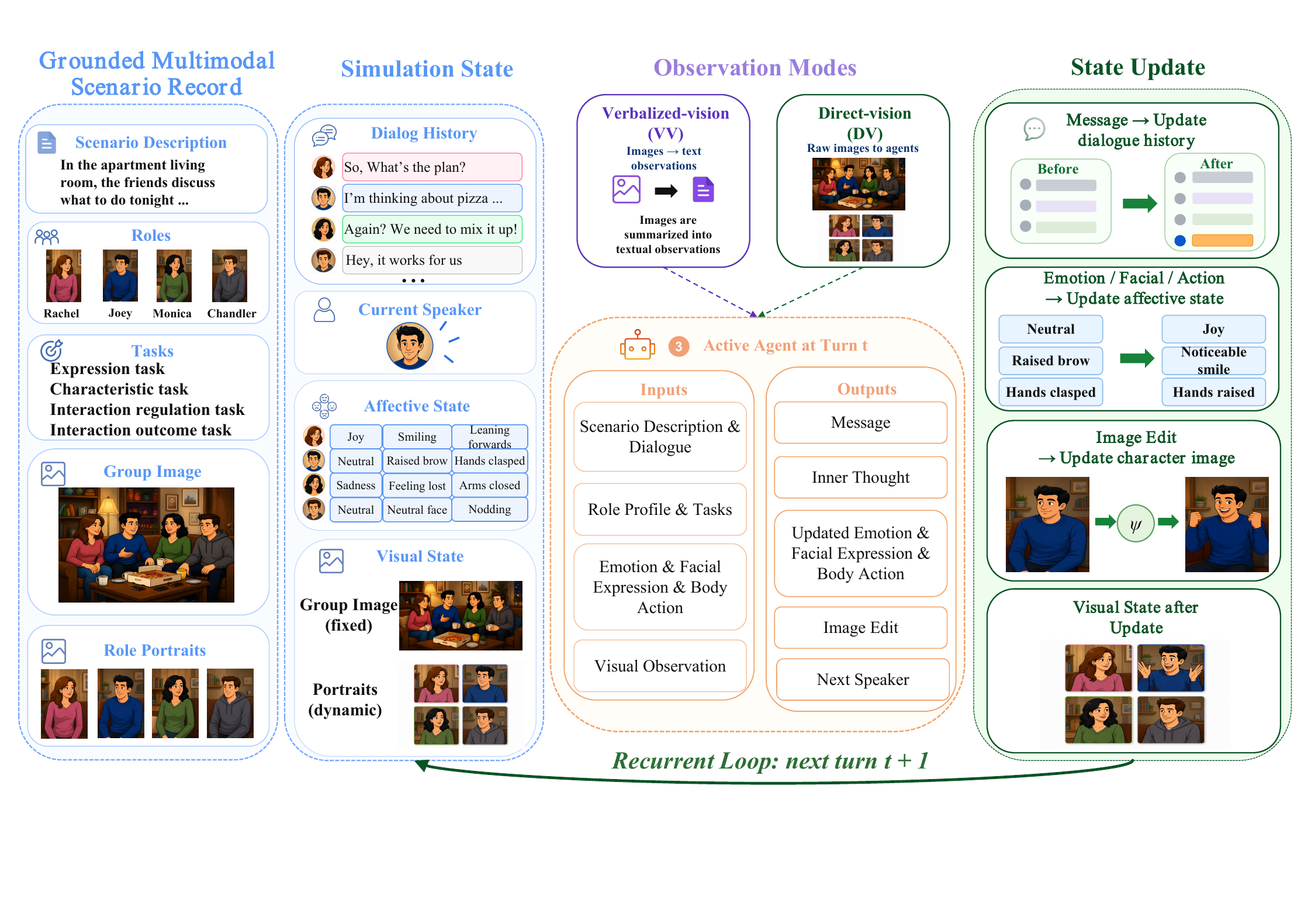}
\caption{
Multimodal social simulation workflow.
A simulation context initializes scenario information, role profiles and tasks, affective states, and visual evidence.
At each turn, the active agent receives \(I_t^m\) under VV or DV and emits a structured output.
The simulator updates dialogue history \(H_t\), current speaker \(a_t\), role affective states \(\mathcal{Z}_t\), and role portraits \(\mathcal{P}_t\), optionally applying \(\Psi\) to edit the active role's portrait before looping to turn \(t{+}1\).
}
    \label{fig:simulation}
\end{figure*}

\subsection{Benchmark Composition and Statistics}
\label{subsec:benchmark-statistics}

Each simulation context in \textsc{\benchmarkname{}} forms a self-contained input package for the simulator described in Section~\ref{sec:simulation}, written as $(G, \mathcal{P}_1, d, \mathcal{C}, \mathcal{Z}_1, 
\mathcal{T})$. The elements $(G, d, \mathcal{C}, \mathcal{T})$ remain fixed throughout the simulation, while role portraits $\mathcal{P}_t$ and affective states $\mathcal{Z}_t$ evolve across turns.

\begin{table}[H]
\centering
\small
\setlength{\tabcolsep}{3.5pt}
\renewcommand{\arraystretch}{1.05}
\begin{tabular}{@{}lr@{\hspace{1.3em}}lr@{\hspace{1.3em}}lr@{}}
\toprule
\multicolumn{2}{c}{\textbf{Conflict level}} &
\multicolumn{2}{c}{\textbf{Dialogue type}} &
\multicolumn{2}{c}{\textbf{Participants}} \\
\cmidrule(r){1-2}\cmidrule(lr){3-4}\cmidrule(l){5-6}
High   & 80 & Persuasion          & 78 & 2 roles & 160 \\
Medium & 80 & Deliberation        & 58 & 3 roles &  59 \\
Low    & 80 & Information-seeking & 56 & 4 roles &  17 \\
       &    & Eristic             & 48 & 5 roles &   4 \\
\bottomrule
\end{tabular}
\caption{
\textsc{\benchmarkname{}} statistics over 240 scenarios. Conflict levels are balanced; dialogue types and participant counts follow the filtered source distribution.
}
\label{tab:benchmark-statistics}
\end{table}

As shown in Table~\ref{tab:benchmark-statistics}, the 240 scenarios are evenly distributed across three conflict intensity levels and cover four dialogue types. Two-role scenarios predominate, with the remainder involving three to five participants. In total, the benchmark yields 585 
role instances and 2{,}340 role-level task instances. Section~\ref{subsec:dialogue-group-factors} analyzes variation across dialogue types and group sizes.

\section{Simulation}
\label{sec:simulation}
For each simulation context, we run the multi-agent simulation under two observation modes. In each run, all participating roles are instantiated simultaneously and act in turn, producing a multi-turn interaction trace for subsequent evaluation (Figure~\ref{fig:simulation}). This section first introduces the two observation modes 
(Section~\ref{sec:observation-modes}), then formalizes the simulation state (Section~\ref{sec:simulation-state}), and describes the per-turn interaction 
protocol (Section~\ref{sec:turn-interaction}).

\subsection{Observation Modes}
\label{sec:observation-modes}

Each simulation is run under one of two observation modes that differ only in how visual evidence is provided to the active agent at each turn.

\begin{itemize}[leftmargin=*]
    \item \textbf{Verbalized-vision (VV).} The acting agent first converts the current group image and role portraits into a textual description of visible facial expressions, body actions, and affective cues, then incorporates this description into its reasoning.
    \item \textbf{Direct-vision (DV).} The agent receives the group image and current role portraits directly as visual inputs, without 
    intermediate verbalization.
\end{itemize}

Both modes provide the same textual context to the active agent but differ in the input form of visual evidence.
These two modes allow us to evaluate agents in social simulations that depend on visual expressions and actions, while reducing the risk of conflating social-interaction failures with failures in direct image understanding. VV makes visual cues explicit through language before reasoning, whereas DV requires agents to perceive these cues directly from images and use them for decision-making. This contrast helps identify whether agents fail because they cannot reason with visual social cues, or because they cannot reliably extract such cues from images.

\subsection{Simulation State}
\label{sec:simulation-state}
For each simulation context in \textsc{\benchmarkname{}}, we distinguish two parts:
\(T_1=(d,\mathcal{C},\mathcal{T},\mathcal{Z}_1)\) contains the textual specification, and 
\(V_1=(G,\mathcal{P}_1)\) contains the visual evidence.
Variables indexed by \(t\) are dynamic quantities updated during simulation, such as \(\mathcal{Z}_t\) and \(\mathcal{P}_t\) denote their values after the simulation has evolved to turn \(t\).
In addition to these context variables, the simulator maintains the dialogue history \(H_t\) (with \(H_1=\varnothing\)) and the current speaker \(a_t\), where \(a_1\) is sampled from the participating roles. 
At turn \(t\), the active speaking agent $a_t$ receives a mode-specific input
\[
I_t^m =
\bigl(d,\ \mathcal{C}^{a_t},\ \mathcal{T}^{a_t},\ H_t,\ \mathcal{Z}_t^{a_t},\ o_t^m\bigr),
\]
where \(\mathcal{C}^{a_t}\), \(\mathcal{T}^{a_t}\), and \(\mathcal{Z}_t^{a_t}\) are the profile, tasks, and current affective state of the active speaker. Note that $m$ represents the observation mode and is chosen from verbalized-vision (VV) and direct-vision (DV). Hence, the input visual observation \(o_t^m\) is either $o_t^{\mathrm{VV}}=\mathrm{verbalize}(V_t)$ or $o_t^{\mathrm{DV}}=V_t$. 
The mode \(m\) is fixed across an entire simulation run.

\begin{table*}[t]
\centering
\resizebox{\textwidth}{!}{
\begin{tabular}{lcccccccccc}
\toprule
\multirow{4}{*}{\textbf{Models}} 
& \multicolumn{10}{c}{\textbf{Role-Level Social Tasks}} \\
\cmidrule(lr){2-11}
& \multicolumn{5}{c}{\textbf{Role Enactment Tasks}} 
& \multicolumn{5}{c}{\textbf{Interaction Management Tasks}} \\
\cmidrule(lr){2-6} \cmidrule(lr){7-11}
& \multicolumn{2}{c}{\textbf{Verbalized-vision}} 
& \multicolumn{2}{c}{\textbf{Direct-vision}} 
& \multirow{2}{*}{\textbf{Avg.}}
& \multicolumn{2}{c}{\textbf{Verbalized-vision}} 
& \multicolumn{2}{c}{\textbf{Direct-vision}} 
& \multirow{2}{*}{\textbf{Avg.}} \\
\cmidrule(lr){2-3} \cmidrule(lr){4-5}
\cmidrule(lr){7-8} \cmidrule(lr){9-10}
& \textbf{Expr.} & \textbf{Char.}
& \textbf{Expr.} & \textbf{Char.}
& 
& \textbf{Int.-Reg.} & \textbf{Int.-Out.}
& \textbf{Int.-Reg.} & \textbf{Int.-Out.}
& \\
\midrule

\textbf{Claude-Sonnet-4.6}     
& \textbf{97.78} & \textbf{97.91}
& \underline{97.52} & \textbf{97.91}
& \textbf{97.78}
& \textbf{71.75} & \textbf{90.64}
& \underline{34.10} & \textbf{88.80}
& \textbf{71.32} \\

\textbf{GLM-4.6V}              
& \underline{97.35} & \textbf{97.91}
& \textbf{97.82} & \underline{97.44}
& \underline{97.63}
& 47.39 & 70.47
& 25.81 & 67.56
& 52.81 \\

\textbf{GPT-5.4}               
& 82.18 & 85.47
& 88.12 & 89.70
& 86.37
& 31.37 & \underline{79.70}
& 25.56 & \underline{82.91}
& 54.89 \\

\textbf{InternVL3.5-241B-A28B} 
& 90.68 & 94.79
& 92.31 & 93.76
& 92.89
& 31.32 & 57.14
& 27.48 & 59.53
& 43.87 \\

\textbf{Qwen3.5-122B-A10B}     
& 97.14 & 96.88
& 96.41 & 95.81
& 96.56
& \underline{61.71} & 77.99
& 30.64 & 77.56
& \underline{61.98} \\

\textbf{Qwen3.5-27B}           
& 96.37 & \underline{97.61}
& 96.71 & 96.54
& 96.81
& 56.28 & 77.35
& 33.97 & 76.15
& 60.94 \\

\textbf{Qwen3.5-9B}            
& 95.98 & 95.43
& 89.15 & 89.62
& 92.55
& 47.35 & 72.86
& \textbf{40.90} & 62.69
& 55.95 \\

\midrule
\textbf{Average}
& 93.93 & 95.14
& 94.01 & 94.40
& 94.37
& 49.60 & 75.16
& 31.21 & 73.60
& 57.39 \\

\bottomrule
\end{tabular}
}
\caption{Performance comparison of simulation models on role enactment tasks and interaction management tasks under two observation modes. For role enactment tasks, Avg.\ averages Expr.\ and Char.\ under both VV and DV. For interaction management tasks, Avg.\ averages Int.-Reg.\ and Int.-Out.\ under both observation modes.}
\label{tab:mode_goal_results}
\end{table*}

\subsection{Turn Interaction and State Update}
\label{sec:turn-interaction}

At each turn, the simulator picks an active speaker \(a_t\) from the previous turn's next-speaker suggestion, subject to a balancing constraint that prevents any single role from dominating. Given the input \(I_t^m\), the agent emits a structured output: a \textbf{public message}, an \textbf{inner thought}, updated \textbf{emotion} / \textbf{facial expression} / \textbf{body action}, a \textbf{next-speaker suggestion}, and an \textbf{image-update decision} \(q_t\in\{0,1\}\).

The state transition appends the message to \(H_{t+1}\), replaces the active role's entry in \(\mathcal{Z}_t\) with the newly generated affective state, and sets \(a_{t+1}\) from the next-speaker suggestion. The portraits are updated as
\[
\mathcal{P}_{t+1} =
\begin{cases}
\Psi(\mathcal{P}_t, a_t, \mathcal{Z}_{t+1}^{a_t}), & q_t=1,\\
\mathcal{P}_t, & q_t=0,
\end{cases}
\]
where \(\mathcal{Z}_{t+1}^{a_t}\) is the updated affective state of the active role and \(\Psi\) edits only that role's portrait. We instantiate \(\Psi\) with LongCat-Image-Edit~\citep{LongCat-Image}. After \(T\) turns, the simulation produces a trace used for evaluation in Section~\ref{sec:evaluation}.

\section{Evaluation}
\label{sec:evaluation}

\subsection{Evaluation Setup}

\paragraph{Implementation Details.} 
Following the simulation procedure in Section~\ref{sec:simulation}, we evaluate whether each simulated role completes its four role-level tasks during the multi-turn interaction. 
Specifically, role enactment tasks and interaction management tasks are judged separately, as they rely on different social evidence. 
For role enactment tasks, we examine whether the dialogue history and inner thoughts of the agent reflect the expression style and conflict characteristic of the assigned role.
For interaction management tasks, we consider the dialogue history, inner thoughts, and all participants' affective states $\mathcal{Z}_t$, since these tasks require the agent to adapt to evolving interaction states and shape the resulting social outcomes.

\paragraph{Metrics.} 
Each role-level task instance is evaluated by judge models and receives one of three final labels: \textit{Achieved}, \textit{Partially Achieved}, or \textit{Not Achieved}. 
We assign these labels numerical weights of 2, 0.5, and 0, respectively. 
For each task dimension, let \(N\), \(N_{\textit{a}}\), and \(N_{\textit{p}}\) denote the total number of evaluated instances and the counts of instances labeled as \textit{Achieved} and \textit{Partially Achieved}, respectively. We compute the normalized task score as
\[
\text{Score}
=
\frac{
2N_{\textit{a}} + 0.5N_{\textit{p}}
}{
2N
}
\times 100.
\]

\paragraph{Models.} We evaluate seven MLLMs from five families: Claude~\citep{anthropic_claude_sonnet_46_model}, GPT~\citep{openai_gpt54}, Qwen~\citep{qwen3.5}, GLM~\citep{vteam2025glm45vglm41vthinkingversatilemultimodal}, and InternVL~\citep{wang2025internvl35advancingopensourcemultimodal}. 
Each role-level task instance is independently labeled by three judge models: Gemini-3.1-pro-preview~\citep{google2026gemini31pro}, GPT-5.4, and Qwen3.5-27B. The final label is determined by majority vote. Appendix~\ref{app:evaluation-protocol} provides more details and inter-judge consistency analysis.
\subsection{Main Results}

\paragraph{Overall Performance}
Table~\ref{tab:mode_goal_results} compares the seven MLLMs on role enactment tasks and interaction management tasks under the two observation modes. Role enactment tasks are close to saturation: all models average above 86, and the overall average reaches 94.37. This indicates that current MLLMs can maintain the specified expression style and conflict characteristic during multi-turn simulation. In contrast, interaction management tasks remain substantially more challenging, with an overall average of 57.39 and a larger cross-model spread. Claude-Sonnet-4.6 performs best with 71.32, while InternVL3.5-241B-A28B obtains 43.87. This suggests that current MLLMs still struggle to regulate interactions and produce intended social outcomes. In particular, VV outperforms DV on Int.-Reg., while the two modes are much closer on Expr., Char., and Int.-Out.

\paragraph{Model-specific Analysis}
Claude-Sonnet-4.6 leads on both task groups. Within the Qwen family, interaction management scores increase monotonically with scale, but this trend does not transfer across families: InternVL3.5-241B-A28B still trails Qwen3.5-9B despite being much larger. GLM-4.6V shows the clearest split between local enactment and interaction management, ranking second on role enactment tasks (97.63) but only sixth on interaction management tasks (52.81). GPT-5.4 performs relatively lower on some role enactment tasks; case inspection suggests that it often chooses milder responses in situations where the task requires assertive, strategic, or confrontational behavior, indicating that safety-oriented response tendencies may reduce completion of such role-specific tasks.

\section{Further Analysis}\label{sec:analysis}

We now look more closely at how visual evidence shapes social behavior on \textsc{\benchmarkname{}}. Section~\ref{subsec:direct-vision} examines why DV underperforms VV on Int.-Reg..
Section~\ref{subsec:vision-dependent-outcomes} shows that the same difficulty extends to Int.-Out.\ when the task itself depends on visible states. Section~\ref{subsec:text-alone-insufficient} then asks the converse question, whether visual evidence is needed at all, by comparing the two vision-enabled modes against a purely text-based interaction setting.
Finally, Section~\ref{subsec:dialogue-group-factors} analyzes how interaction-management performance varies with dialogue type and the number of participating roles.

\subsection{Direct Vision Is Not Enough}
\label{subsec:direct-vision}

\begin{figure}[t]
    \centering
    \includegraphics[width=0.9\linewidth,
    trim=0 5pt 0 20pt,
    clip]{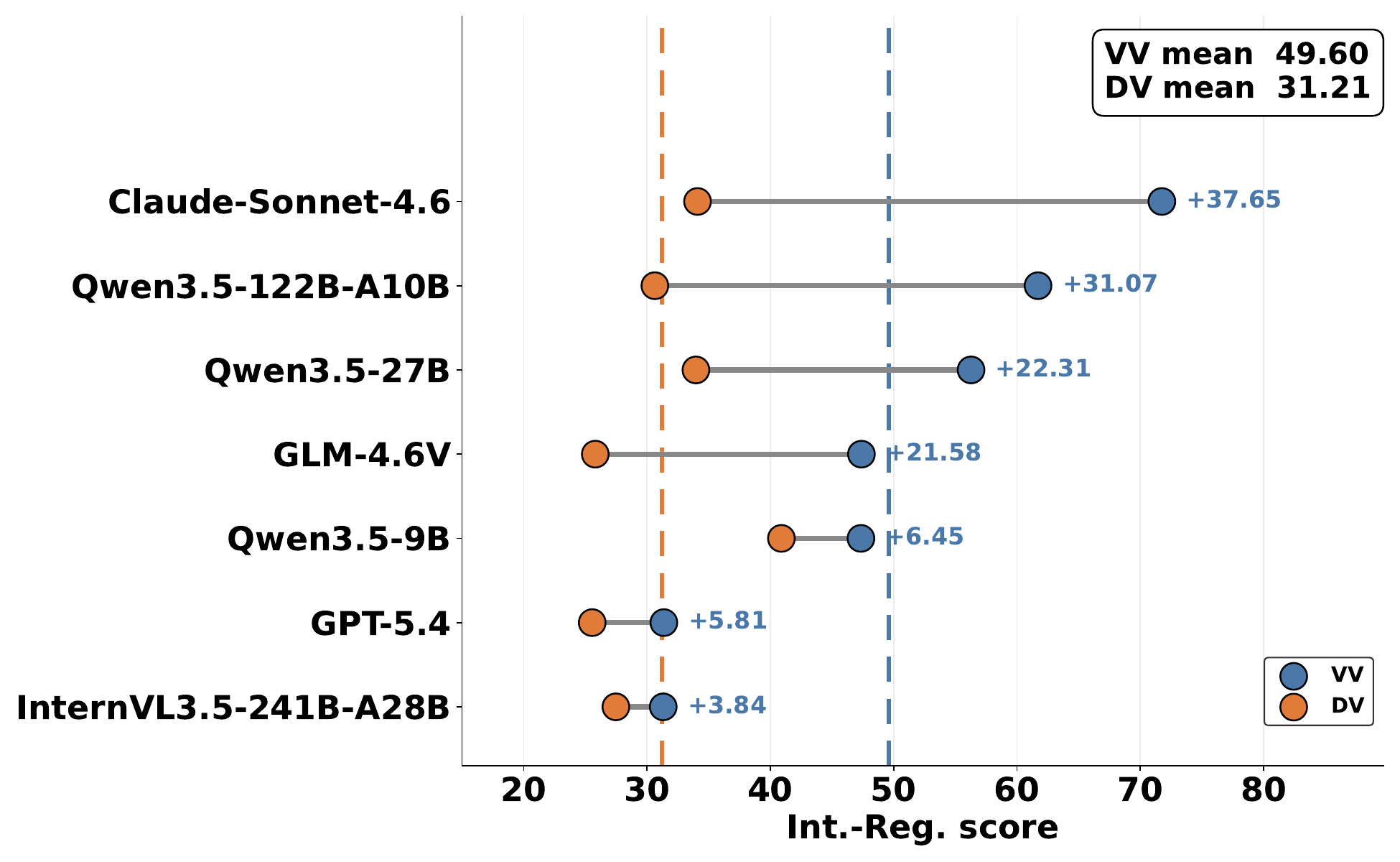}
    \caption{
    Per-model Int.-Reg.\ scores under VV and DV, sorted by VV$-$DV 
  gap. Dotted lines mark VV and DV means (49.60 vs.\ 31.21). All seven models score higher under VV, with the largest gain observed for Claude-Sonnet-4.6.
    }
    \label{fig:int_reg_vv_dv_gap}
\end{figure}

As shown in Table~\ref{tab:mode_goal_results} and Figure~\ref{fig:int_reg_vv_dv_gap}, Int.-Reg.\ exhibits a much larger VV--DV gap than the other task dimensions. Averaged across models, the score increases from 31.21 under DV to 49.60 under VV, yielding an 18.39-point gain, while the gaps on the other three dimensions remain below two points. Since Int.-Reg.\ requires agents to track others' verbal and non-verbal states and adapt their own responses accordingly, this gap raises the possibility that models do not always use raw visual input effectively for interaction regulation. 
We further rule out a trivial verbosity explanation: VV and DV produce nearly the same amount of public dialogue per role, yet VV improves Int.-Reg.\ efficiency by about 54\% over DV (Appendix~\ref{app:communication-efficiency}). Thus, the VV advantage is unlikely to come from longer conversations, but rather from visual cues being more usable for regulation.

We next examine whether DV failures mainly stem from visual recognition. If models fail under DV simply because they cannot perceive the relevant facial expressions or postures, then their textual visual descriptions in VV should also be poor. We therefore manually score VV descriptions for 10 scenarios across four models, yielding 640 ratings on a 1--3 scale. Table~\ref{tab:human_check_main} summarizes the results, with the full distribution reported in Appendix~\ref{app:human-check}. The ratings suggest that the VV--DV gap is not a single perception failure. Claude-Sonnet-4.6 achieves near-perfect description quality, and shows the largest Int.-Reg.\ gain from VV, indicating a perception-to-decision integration bottleneck: the model can recover relevant social cues when explicitly prompted, but does not use them as reliably when they enter as raw images. For smaller models, especially on group images, visual recognition and decision integration may both contribute. Thus, direct visual access alone is not sufficient.

\begin{table}[t]
\centering
\small
\setlength{\tabcolsep}{3.5pt}
\renewcommand{\arraystretch}{1.12}
\resizebox{\linewidth}{!}{
\begin{tabular}{lccc}
\toprule
\textbf{Agent} & \textbf{Group / Portrait} & \textbf{Mean} & \textbf{Int.-Reg.\ $\Delta$} \\
\midrule
Claude-Sonnet-4.6 & 2.80 / 2.95 & \textbf{2.94} & \textbf{+37.65} \\
Qwen3.5-122B-A10B & 1.80 / 2.77 & 2.71 & +31.07 \\
Qwen3.5-27B & 1.40 / 2.65 & 2.58 & +22.31 \\
Qwen3.5-9B & 1.20 / 2.15 & 2.09 & +6.45 \\
\bottomrule
\end{tabular}
}
\caption{
Human check of visual descriptions under VV and the corresponding Int.-Reg.\ gain. Ratings are on a 1--3 scale; \(\Delta\) is VV minus DV.
}
\label{tab:human_check_main}
\end{table}
\subsection{Strongly Vision-Dependent Outcomes Are Challenging}
\label{subsec:vision-dependent-outcomes}

The previous analysis shows that models have difficulty converting visual cues into interaction regulation.  We next examine whether Int.-Out.\  that depend on visual evidence are also harder for models. Int.-Out.\ evaluates whether the model eventually achieves a concrete social result involving at least one other participant. Since some Int.-Out.\ outcomes depend on visually observable social states, we split Int.-Out.\ instances by whether successful completion requires visual evidence. Strongly vision-dependent tasks require concrete cues such as facial expression, posture, gaze or salient emotional shifts, whereas weakly vision-dependent tasks can be completed mainly through dialogue, reasoning, or interaction structure. The split is generated by GPT-5.4 and manually validated, resulting in 85 strongly vision-dependent tasks, or 14.53\% of all Int.-Out.\ . Details are provided in Appendix~\ref{app:vision-intensive-classification}.

\begin{figure}[t]
    \centering
    \includegraphics[
        width=0.9\linewidth,
    trim=0 10pt 0 0pt,
    clip
    ]{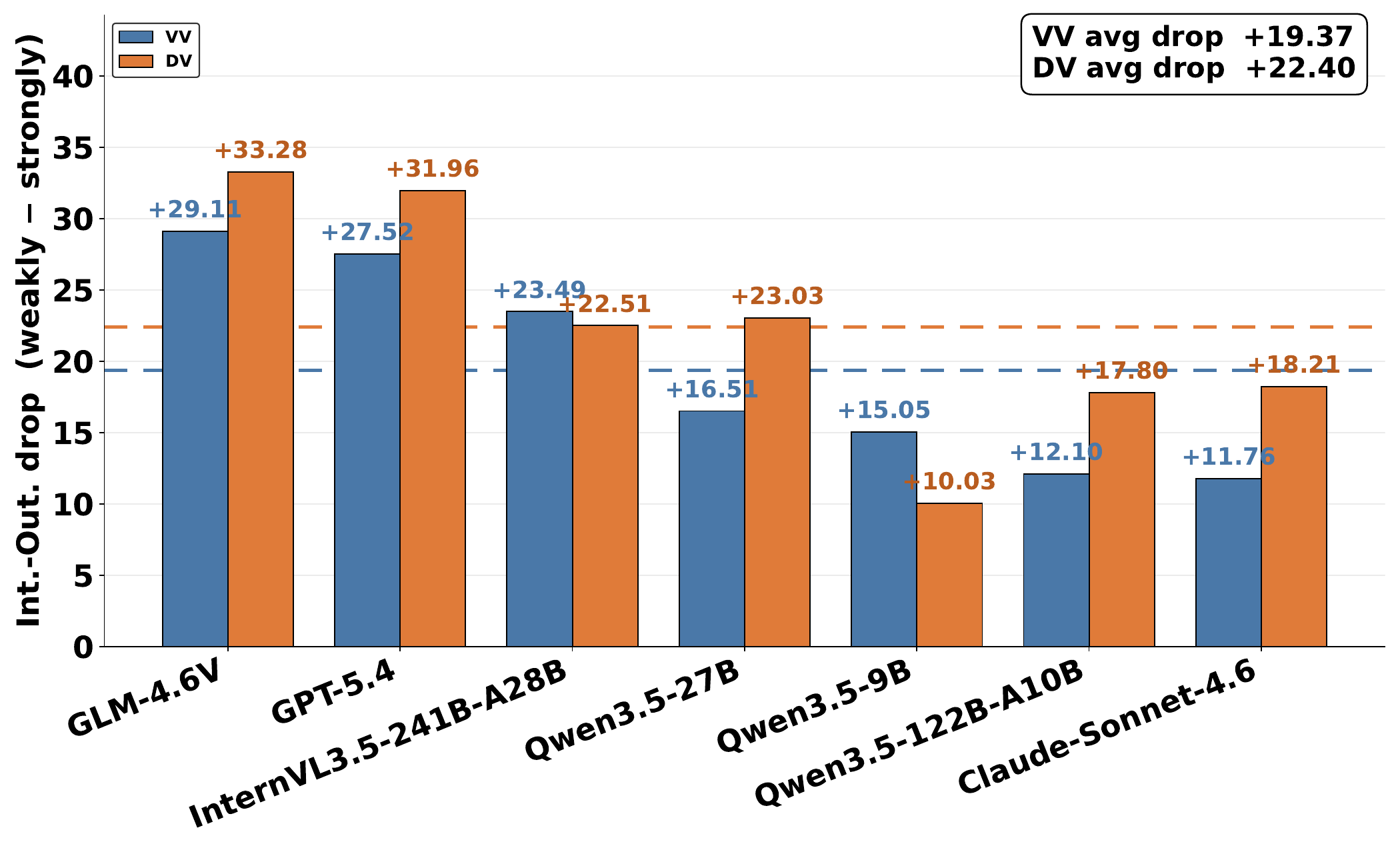}
    \caption{
Per-model Int.-Out.\ score drop from weakly to strongly vision-dependent tasks, under VV and DV. Dashed lines mark the average drop in each mode.
}
    \label{fig:vision_intensive}
\end{figure}

Figure~\ref{fig:vision_intensive} shows that all models perform worse on strongly vision-dependent Int.-Out.\ . Averaged across models and modes, the score drops from 77.42 on weakly vision-dependent tasks to 56.53 on strongly vision-dependent tasks. The drop appears under both VV and DV, indicating that vision-dependent outcomes remain difficult even when models are given access to visual information, either through explicit descriptions or direct images. Int.-Reg.\ evaluates local adaptation to others' current states, while vision-dependent Int.-Out.\ requires such adaptation to accumulate into a concrete social result. Taken together with the Int.-Reg.\ gap in Section~\ref{subsec:direct-vision}, this result suggests that models still struggle to translate visual cues into effective interaction decisions, both when adjusting to others' immediate states and when pursuing downstream social outcomes.

\subsection{Text Alone Is Insufficient}
\label{subsec:text-alone-insufficient}

We next test whether visual evidence is necessary by comparing VV and DV with a no-vision baseline, \textbf{Text-only}, where agents receive only textual context and no group image or role portraits.

\begin{figure}[t]
    \centering
    \includegraphics[
        width=\linewidth,
    trim=0 10pt 0 5pt,
    clip
    ]{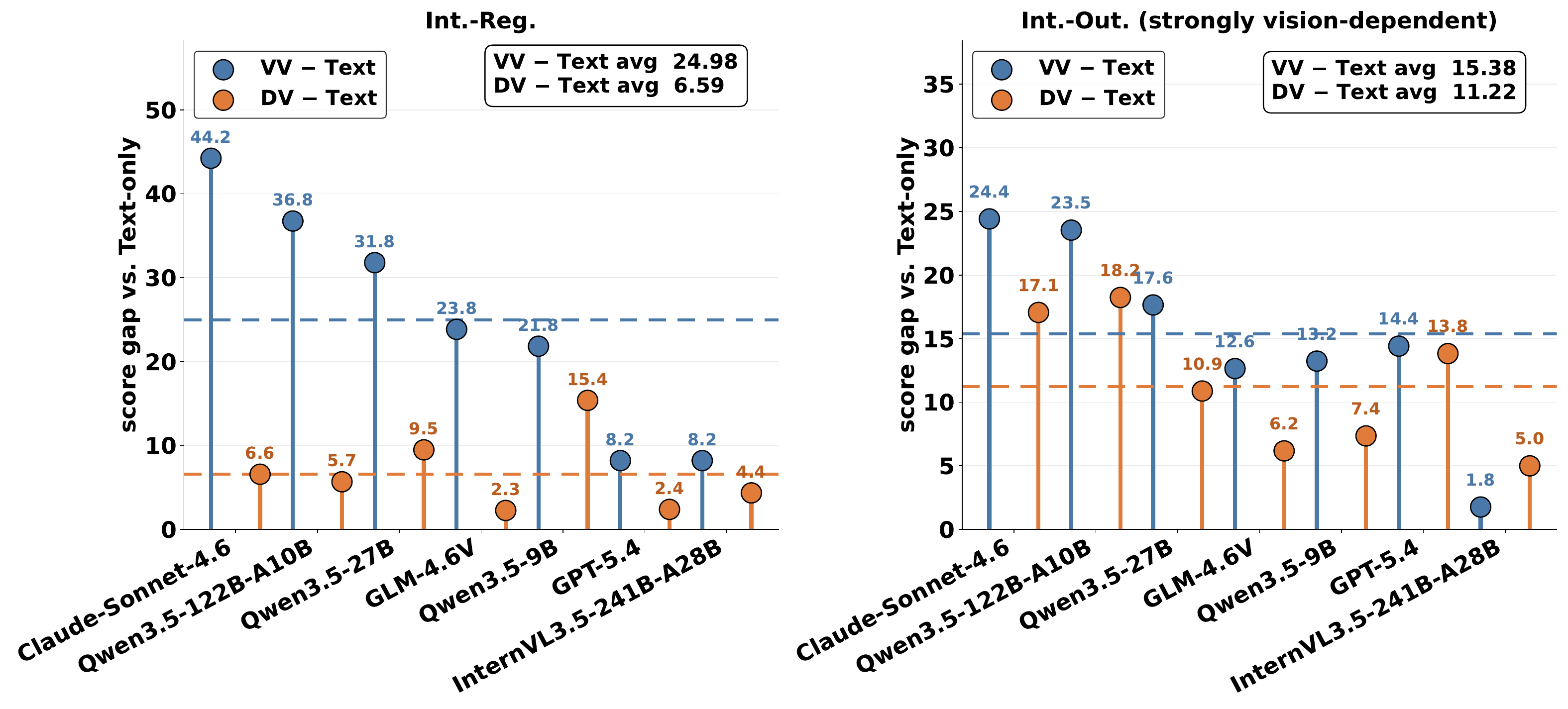}
    \caption{Score gaps relative to the Text-only baseline. Dashed lines mark the average gap in each mode.
    }
    \label{fig:vision_necessity_gap}
\end{figure}

We focus on Int.-Reg.\ tasks and the strongly vision-dependent Int.-Out.\ . As shown in Figure~\ref{fig:vision_necessity_gap}, both VV and DV consistently outperform Text-only across models, showing that visual evidence provides social information beyond dialogue and textual context. These gains indicate that visual cues are beneficial for social tasks that depend on observable affective and behavioral states. Taken together with the VV--DV gap in Section~\ref{subsec:direct-vision} and the drop on strongly vision-dependent Int.-Out.\ in Section~\ref{subsec:vision-dependent-outcomes}, these results indicate that visual information is beneficial for vision-dependent social interaction, but models still struggle to convert visual cues into effective interaction decisions.

\subsection{Interaction Management Tasks Vary Across Scenarios}
\label{subsec:dialogue-group-factors}

The previous sections show that visual evidence is useful for dynamic social interaction, but that models still struggle to translate visual cues into effective interaction decisions. We further examine whether performance on Interaction Management Tasks varies across scenarios, focusing on scenario-level dialogue type and the number of participating characters. Scores follow the same model-balanced averaging protocol as Table~\ref{tab:mode_goal_results}. The analysis is descriptive: dialogue type and group size are naturally coupled with scenario content, so the goal is to identify useful performance patterns rather than to isolate causal effects.

\begin{table}[t]
\centering

\small
\setlength{\tabcolsep}{4pt}
\renewcommand{\arraystretch}{1.12}
\begin{tabular}{lrrrrr}
\toprule
\textbf{Dialogue Type} & \textbf{Scen.} & \multicolumn{2}{c}{\textbf{Int.-Reg.}} & \multicolumn{2}{c}{\textbf{Int.-Out.}} \\
\cmidrule(lr){3-4}\cmidrule(lr){5-6}
& & \textbf{VV} & \textbf{DV} & \textbf{VV} & \textbf{DV} \\
\midrule
Information-seeking & 56 & 41.83 & 27.46 & \textbf{79.44} & \textbf{78.27} \\
Deliberation & 58 & 51.02 & 28.38 & 72.64 & 70.48 \\
Persuasion & 78 & 51.48 & 34.05 & 74.27 & 73.11 \\
Eristic & 48 & \textbf{53.32} & \textbf{34.37} & 74.89 & 73.05 \\
\bottomrule
\end{tabular}
\caption{
Task completion by dialogue type, restricted to the two dynamic interaction
dimensions. The largest type-level variation appears on \textit{interaction
regulation}, where information-seeking scenarios are more difficult than
eristic, persuasion, and deliberation scenarios under both observation modes.
}
\label{tab:dialogue-type-performance}
\end{table}

Table~\ref{tab:dialogue-type-performance} shows that dialogue type mainly affects the Int.-Reg.  of Interaction Management Tasks. Information-seeking scenarios obtain the lowest Int.-Reg.
scores under both Verbalized-vision and Direct-vision, suggesting that
asymmetrical question-answer roles make it harder for agents to turn observed partner states into adaptive conversational moves. In contrast, eristic and persuasion scenarios obtain higher Int.-Reg. scores, likely because their conflictive goals make social stances and response pressures more explicit. By comparison, Int.-Out. varies less across dialogue types, suggesting that dialogue type has a stronger effect on local interaction regulation than on eventual outcome completion.

\begin{table}[t]
\centering

\small
\setlength{\tabcolsep}{4pt}
\renewcommand{\arraystretch}{1.12}
\begin{tabular}{lrrrrr}
\toprule
\textbf{Group Size} & \textbf{Scen.} & \multicolumn{2}{c}{\textbf{Int.-Reg.}} & \multicolumn{2}{c}{\textbf{Int.-Out.}} \\
\cmidrule(lr){3-4}\cmidrule(lr){5-6}
& & \textbf{VV} & \textbf{DV} & \textbf{VV} & \textbf{DV} \\
\midrule
2 characters & 160 & \textbf{53.67} & 31.60 & 75.13 & 74.52 \\
3 characters & 59 & 45.04 & 30.75 & \textbf{79.04} & \textbf{75.30} \\
4 characters & 17 & 47.79 & \textbf{32.20} & 67.96 & 66.91 \\
5 characters & 4 & 30.89 & 25.71 & 65.89 & 66.61 \\
\bottomrule
\end{tabular}
\caption{
Task completion by number of participating characters, restricted to the two
dynamic interaction dimensions. Int.-Reg.\ drops as more participants must be
tracked, especially under Verbalized-vision; Int.-Out.\ also weakens for the
largest groups.
}
\label{tab:group-size-performance}
\end{table}

Table~\ref{tab:group-size-performance} further shows that the number of participating characters affects Interaction Management Tasks. The clearest pattern again appears on Int.-Reg.: two-character scenarios achieve the highest Verbalized-vision Int.-Reg. score, while
five-character scenarios are substantially lower. This suggests that as more participants enter the interaction, agents must track more visible states, infer more interpersonal dependencies, and choose responses that remain coherent across multiple partners. Int.-Out. also weakens for the largest groups, although the five-character subset is small. Overall, these results indicate that Interaction Management Tasks are shaped not only by the availability and format of visual evidence, but also by the dialogue structure and the number of participants involved in the social interaction.
\section{Related Work}\label{sec:related-work}

\paragraph{Social agents and simulation.}
Social-agent benchmarks evaluate goal-driven interaction such as cooperation, conflict, negotiation, and persuasion~\citep{zhou2024sotopia,mou-etal-2025-agentsense}, while role-playing benchmarks focus on persona consistency, character enactment, and group dynamics~\citep{chen2024socialbenchsocialityevaluationroleplaying,tu2024characterevalchinesebenchmarkroleplaying}. 
Generative-agent systems further simulate memory-augmented societies and long-horizon social behavior~\citep{park2023generativeagentsinteractivesimulacra,wang2024sotopiapiinteractivelearningsocially,piao2026agentsocietylargescalesimulationllmdriven}. 
These works provide important testbeds for text-based social interaction, but rarely examine how vision evidence enters agents' turn-level decisions. 

\paragraph{Multimodal social understanding.}
Multimodal social datasets evaluate affect and relational inference from text, audio, image, or video~\citep{poria2019meldmultimodalmultipartydataset,8953344,vicol2018moviegraphsunderstandinghumancentricsituations,park2020visualcomet}, and recent benchmarks extend to multimodal theory-of-mind and emotional shifts. In contrast, \textsc{\benchmarkname{}} focuses on visual social intelligence in simulation, evaluating both local role enactment and interaction management.
\section{Conclusion}\label{sec:conclusion}

We introduced \textsc{\benchmarkname{}}, a benchmark evaluating visual social intelligence in multimodal social simulation. By combining aligned textual and visual evidence with role-level social tasks, \textsc{\benchmarkname{}} evaluates not only local role enactment but also interaction regulation and outcome formation. 
Experiments on seven recent MLLMs show that agents often maintain assigned expression style and conflict characteristic, but struggle to use visible social cues for regulation and outcome formation, especially when those cues are provided as raw images rather than verbalized observations.

\section*{Limitations}
Our scenarios are constructed from a single Western situational comedy (\textit{Friends}), which              
underrepresents cross-cultural conventions, professional settings, and non-English social norms, so the
generality of the perception-to-decision gap we observe remains open. Each scene is represented by a group    
image and per-role portraits at fixed turns rather than continuous video, leaving temporal cues such as gaze
shifts and micro-expressions only partially captured. Finally, task completion is scored by an LLM-as-judge
ensemble with a three-label rubric: although we use majority voting across three judges, judge models may share biases with the agents
they evaluate, and the coarse rubric cannot capture finer distinctions of social appropriateness.

\section*{Ethical Considerations}

\textsc{\benchmarkname{}} is built from the TV series \textit{Friends}, a source also used by prior dialogue and multimodal datasets for emotion recognition, question answering, relation extraction, and multi-party conversation understanding~\citep{poria-etal-2019-meld,yang-choi-2019-friendsqa,yu2020dialoguebasedrelationextraction,wang2024friendsmmcdatasetmultimodalmultiparty}. 
However, the source videos, frames, and character appearances are copyrighted media. 
Unless explicitly permitted, we do not redistribute raw episode videos or original frames; releases should focus on annotations, metadata, prompts, and extraction code that require users to access source media through legitimate channels.

We additionally use GPT-image-1~\citep{openai_gptimage_api} to generate synthetic role portraits from source visual evidence for controlled simulation. 
These portraits are not authentic depictions of the original actors, nor claims about their real emotions, traits, or intentions. 
To reduce the risk of misleading or derogatory portrayals, generated portraits are treated as synthetic research artifacts, clearly marked as such, and not intended to mock, defame, or harm depicted characters or actors.

The benchmark contains affective cues, conflict, persuasion, avoidance, deception, and other socially sensitive behaviors. 
It is intended for diagnostic research on multimodal agents, not for real-world emotion inference or high-stakes social, clinical, legal, or workplace decision-making.

\bibliography{references}

\begin{thebibliography}{43}
\providecommand{\natexlab}[1]{#1}

\bibitem[{Adam and Gaudou(2016)}]{adam2016bdi}
Carole Adam and Benoit Gaudou. 2016.
\newblock \href {https://doi.org/10.1017/S0269888916000096} {Bdi agents in social simulations: a survey}.
\newblock \emph{The Knowledge Engineering Review}, 31:207--238.

\bibitem[{{Anthropic}(2026)}]{anthropic_claude_sonnet_46_model}
{Anthropic}. 2026.
\newblock Claude sonnet 4.6.
\newblock \url{https://www.anthropic.com/claude/sonnet}.
\newblock Accessed: 2026-05-23.

\bibitem[{Bratman(1987)}]{bratman1987intention}
M.~Bratman. 1987.
\newblock \href {https://books.google.com.sg/books?id=I0nuAAAAMAAJ} {\emph{Intention, Plans, and Practical Reason}}.
\newblock Harvard University Press.

\bibitem[{Budagam(2024)}]{Budagam_ROLEBENCH-_A_Role_2024}
Devichand Budagam. 2024.
\newblock \href {https://github.com/devichand579/ROLEBENCH} {{ROLEBENCH- A Role Prompting Benchmark}}.

\bibitem[{Chen et~al.(2024)Chen, Chen, Yan, Xu, Gao, Shen, Quan, Li, Zhang, Huang, and Zhou}]{chen2024socialbenchsocialityevaluationroleplaying}
Hongzhan Chen, Hehong Chen, Ming Yan, Wenshen Xu, Xing Gao, Weizhou Shen, Xiaojun Quan, Chenliang Li, Ji~Zhang, Fei Huang, and Jingren Zhou. 2024.
\newblock \href {https://arxiv.org/abs/2403.13679} {Socialbench: Sociality evaluation of role-playing conversational agents}.
\newblock \emph{Preprint}, arXiv:2403.13679.

\bibitem[{Chen et~al.(2025)Chen, Xiao, Zhang, Luo, Lian, and Liu}]{chen2025m3embeddingmultilingualitymultifunctionalitymultigranularity}
Jianlv Chen, Shitao Xiao, Peitian Zhang, Kun Luo, Defu Lian, and Zheng Liu. 2025.
\newblock \href {https://arxiv.org/abs/2402.03216} {M3-embedding: Multi-linguality, multi-functionality, multi-granularity text embeddings through self-knowledge distillation}.
\newblock \emph{Preprint}, arXiv:2402.03216.

\bibitem[{{Google DeepMind}(2026)}]{google2026gemini31pro}
{Google DeepMind}. 2026.
\newblock Gemini 3.1 pro model card.
\newblock \url{https://deepmind.google/models/model-cards/gemini-3-1-pro/}.
\newblock Accessed: 2026-05-20.

\bibitem[{Hess(2016)}]{NonverbalCommunicationHess}
U.~Hess. 2016.
\newblock \href {https://doi.org/10.1016/B978-0-12-397045-9.00218-4} {Nonverbal communication}.
\newblock In Howard~S. Friedman, editor, \emph{Encyclopedia of Mental Health (Second Edition)}, second edition edition, pages 208--218. Academic Press, Oxford.

\bibitem[{Jang et~al.(2023)Jang, Yoon, Choi, Ong, and Kim}]{jang2023structured}
Minsu Jang, Youngwoo Yoon, Jaewoo Choi, Hyobin Ong, and Jaehong Kim. 2023.
\newblock \href {https://doi.org/10.1145/3623809.3623930} {A structured prompting based on belief-desire-intention model for proactive and explainable task planning}.
\newblock In \emph{Proceedings of the 11th International Conference on Human-Agent Interaction}, HAI '23, page 375–377, New York, NY, USA. Association for Computing Machinery.

\bibitem[{Killough et~al.(2016)Killough, Bauters, McAreavey, Liu, and Hong}]{killough2016riskaware}
Ronan Killough, Kim Bauters, Kevin McAreavey, Weiru Liu, and Jun Hong. 2016.
\newblock \href {https://doi.org/10.5220/0005703103220329} {Risk-aware planning in bdi agents}.
\newblock In \emph{Proceedings of the 8th International Conference on Agents and Artificial Intelligence - Volume 2: ICAART}, pages 322--329. INSTICC, SciTePress.

\bibitem[{Kwon et~al.(2023)Kwon, Li, Zhuang, Sheng, Zheng, Yu, Gonzalez, Zhang, and Stoica}]{kwon2023efficientmemorymanagementlarge}
Woosuk Kwon, Zhuohan Li, Siyuan Zhuang, Ying Sheng, Lianmin Zheng, Cody~Hao Yu, Joseph~E. Gonzalez, Hao Zhang, and Ion Stoica. 2023.
\newblock \href {https://arxiv.org/abs/2309.06180} {Efficient memory management for large language model serving with pagedattention}.
\newblock \emph{Preprint}, arXiv:2309.06180.

\bibitem[{McCornack(1992)}]{information1992steven}
Steven~A. McCornack. 1992.
\newblock \href {https://doi.org/10.1080/03637759209376245} {Information manipulation theory}.
\newblock \emph{Communication Monographs}, 59(1):1--16.

\bibitem[{Mou et~al.(2025)Mou, Liang, Lin, Zhang, Liu, Yang, Ye, Chen, Kuang, Huang, and Wei}]{mou-etal-2025-agentsense}
Xinyi Mou, Jingcong Liang, Jiayu Lin, Xinnong Zhang, Xiawei Liu, Shiyue Yang, Rong Ye, Lei Chen, Haoyu Kuang, Xuanjing Huang, and Zhongyu Wei. 2025.
\newblock \href {https://doi.org/10.18653/v1/2025.naacl-long.257} {{A}gent{S}ense: Benchmarking social intelligence of language agents through interactive scenarios}.
\newblock In \emph{Proceedings of the 2025 Conference of the Nations of the Americas Chapter of the Association for Computational Linguistics: Human Language Technologies (Volume 1: Long Papers)}, pages 4975--5001, Albuquerque, New Mexico. Association for Computational Linguistics.

\bibitem[{{OpenAI}(2026)}]{openai_gpt54}
{OpenAI}. 2026.
\newblock Gpt-5.4 model.
\newblock \url{https://developers.openai.com/api/docs/models/gpt-5.4}.
\newblock Accessed: 2026-05-20.

\bibitem[{{OpenAI}(n.d.{\natexlab{a}})}]{openai_gpt5mini_api}
{OpenAI}. n.d.{\natexlab{a}}.
\newblock {GPT-5 mini Model}.
\newblock \url{https://developers.openai.com/api/docs/models/gpt-5-mini}.
\newblock Accessed: 2026-05-24.

\bibitem[{{OpenAI}(n.d.{\natexlab{b}})}]{openai_gptimage_api}
{OpenAI}. n.d.{\natexlab{b}}.
\newblock {GPT Image 1 Model}.
\newblock \url{https://developers.openai.com/api/docs/models/gpt-image-1}.
\newblock Accessed: 2026-05-24.

\bibitem[{Park et~al.(2020)Park, Bhagavatula, Mottaghi, Farhadi, and Choi}]{park2020visualcomet}
Jae~Sung Park, Chandra Bhagavatula, Roozbeh Mottaghi, Ali Farhadi, and Yejin Choi. 2020.
\newblock \href {https://doi.org/10.1007/978-3-030-58558-7_30} {Visualcomet: Reasoning about the dynamic context of a still image}.
\newblock In \emph{Computer Vision -- ECCV 2020}, pages 508--524, Cham. Springer International Publishing.

\bibitem[{Park et~al.(2023)Park, O'Brien, Cai, Morris, Liang, and Bernstein}]{park2023generativeagentsinteractivesimulacra}
Joon~Sung Park, Joseph~C. O'Brien, Carrie~J. Cai, Meredith~Ringel Morris, Percy Liang, and Michael~S. Bernstein. 2023.
\newblock \href {https://arxiv.org/abs/2304.03442} {Generative agents: Interactive simulacra of human behavior}.
\newblock \emph{Preprint}, arXiv:2304.03442.

\bibitem[{Piao et~al.(2026)Piao, Yan, Zhang, Li, Yan, Lan, Lu, Zheng, Wang, Zhou, Gao, Xu, Zhang, Rong, Su, and Li}]{piao2026agentsocietylargescalesimulationllmdriven}
Jinghua Piao, Yuwei Yan, Jun Zhang, Nian Li, Junbo Yan, Xiaochong Lan, Zhihong Lu, Zhiheng Zheng, Jing~Yi Wang, Di~Zhou, Chen Gao, Fengli Xu, Fang Zhang, Ke~Rong, Jun Su, and Yong Li. 2026.
\newblock \href {https://arxiv.org/abs/2502.08691} {Agentsociety: Large-scale simulation of llm-driven generative agents advances understanding of human behaviors and society}.
\newblock \emph{Preprint}, arXiv:2502.08691.

\bibitem[{Poria et~al.(2019{\natexlab{a}})Poria, Hazarika, Majumder, Naik, Cambria, and Mihalcea}]{poria2019meldmultimodalmultipartydataset}
Soujanya Poria, Devamanyu Hazarika, Navonil Majumder, Gautam Naik, Erik Cambria, and Rada Mihalcea. 2019{\natexlab{a}}.
\newblock \href {https://arxiv.org/abs/1810.02508} {Meld: A multimodal multi-party dataset for emotion recognition in conversations}.
\newblock \emph{Preprint}, arXiv:1810.02508.

\bibitem[{Poria et~al.(2019{\natexlab{b}})Poria, Hazarika, Majumder, Naik, Cambria, and Mihalcea}]{poria-etal-2019-meld}
Soujanya Poria, Devamanyu Hazarika, Navonil Majumder, Gautam Naik, Erik Cambria, and Rada Mihalcea. 2019{\natexlab{b}}.
\newblock \href {https://doi.org/10.18653/v1/P19-1050} {{MELD}: A multimodal multi-party dataset for emotion recognition in conversations}.
\newblock In \emph{Proceedings of the 57th Annual Meeting of the Association for Computational Linguistics}, pages 527--536, Florence, Italy. Association for Computational Linguistics.

\bibitem[{{Qwen Team}(2026)}]{qwen3.5}
{Qwen Team}. 2026.
\newblock \href {https://qwen.ai/blog?id=qwen3.5} {{Qwen3.5}: Towards native multimodal agents}.

\bibitem[{Radford et~al.(2021)Radford, Kim, Hallacy, Ramesh, Goh, Agarwal, Sastry, Askell, Mishkin, Clark, Krueger, and Sutskever}]{radford2021learningtransferablevisualmodels}
Alec Radford, Jong~Wook Kim, Chris Hallacy, Aditya Ramesh, Gabriel Goh, Sandhini Agarwal, Girish Sastry, Amanda Askell, Pamela Mishkin, Jack Clark, Gretchen Krueger, and Ilya Sutskever. 2021.
\newblock \href {https://arxiv.org/abs/2103.00020} {Learning transferable visual models from natural language supervision}.
\newblock \emph{Preprint}, arXiv:2103.00020.

\bibitem[{Rao and George(1995)}]{rao1995bdi}
Anand~S. Rao and Michael~P. George. 1995.
\newblock \href {https://api.semanticscholar.org/CorpusID:269838374} {Bdi agents: from theory to practice}.

\bibitem[{Redmon et~al.(2016)Redmon, Divvala, Girshick, and Farhadi}]{redmon2016lookonceunifiedrealtime}
Joseph Redmon, Santosh Divvala, Ross Girshick, and Ali Farhadi. 2016.
\newblock \href {https://arxiv.org/abs/1506.02640} {You only look once: Unified, real-time object detection}.
\newblock \emph{Preprint}, arXiv:1506.02640.

\bibitem[{Ren et~al.(2022)Ren, Lattas, Gecer, Deng, Ma, Yang, and Zafeiriou}]{ren2022facialgeometricrecoveryimplicit}
Xingyu Ren, Alexandros Lattas, Baris Gecer, Jiankang Deng, Chao Ma, Xiaokang Yang, and Stefanos Zafeiriou. 2022.
\newblock \href {https://arxiv.org/abs/2203.09692} {Facial geometric detail recovery via implicit representation}.
\newblock \emph{Preprint}, arXiv:2203.09692.

\bibitem[{Rimé(1982)}]{elimination}
Bernard Rimé. 1982.
\newblock \href {https://doi.org/10.1002/ejsp.2420120201} {The elimination of visible behaviour from social interactions: Effects on verbal, nonverbal and interpersonal variables}.
\newblock \emph{European Journal of Social Psychology}, 12(2):113--129.

\bibitem[{Team et~al.(2025{\natexlab{a}})Team, Ma, Tan, Huang, Wu, He, Gao, Xiao, Wei, Ma, Cai, Guan, and Hu}]{LongCat-Image}
Meituan~LongCat Team, Hanghang Ma, Haoxian Tan, Jiale Huang, Junqiang Wu, Jun-Yan He, Lishuai Gao, Songlin Xiao, Xiaoming Wei, Xiaoqi Ma, Xunliang Cai, Yayong Guan, and Jie Hu. 2025{\natexlab{a}}.
\newblock \href {https://arxiv.org/abs/2512.07584} {Longcat-image technical report}.
\newblock \emph{Preprint}, arXiv:2512.07584.

\bibitem[{Team et~al.(2025{\natexlab{b}})Team, Hong, Yu, Gu, Wang, Gan, Tang, Cheng, Qi, Ji, Pan, Duan, Wang, Wang, Cheng, He, Su, Yang, Pan, Zeng, Wang, Chen, Shi, Pang, Zhang, Yin, Yang, Chen, Xu, Zhu, Chen, Chen, Chen, Lin, Wang, Chen, Lei, Gong, Pan, Liu, Xu, Zhang, Zheng, Yang, Zhong, Huang, Zhao, Xue, Tu, Meng, Zhang, Luo, Hao, Tong, Li, Jia, Liu, Zhang, Lyu, Fan, Huang, Wang, Xue, Wang, Wang, An, Du, Shi, Huang, Niu, Wang, Yue, Li, Zhang, Wang, Wang, Zhang, Xue, Hou, Du, Wang, Zhang, Liu, Xu, Li, Huang, Dong, and Tang}]{vteam2025glm45vglm41vthinkingversatilemultimodal}
V~Team, Wenyi Hong, Wenmeng Yu, Xiaotao Gu, Guo Wang, Guobing Gan, Haomiao Tang, Jiale Cheng, Ji~Qi, Junhui Ji, Lihang Pan, Shuaiqi Duan, Weihan Wang, Yan Wang, Yean Cheng, Zehai He, Zhe Su, Zhen Yang, Ziyang Pan, and 69 others. 2025{\natexlab{b}}.
\newblock \href {https://arxiv.org/abs/2507.01006} {Glm-4.5v and glm-4.1v-thinking: Towards versatile multimodal reasoning with scalable reinforcement learning}.
\newblock \emph{Preprint}, arXiv:2507.01006.

\bibitem[{Thomas(1992)}]{thomas1992conflict}
Kenneth~W. Thomas. 1992.
\newblock \href {http://www.jstor.org/stable/2488472} {Conflict and conflict management: Reflections and update}.
\newblock \emph{Journal of Organizational Behavior}, 13(3):265--274.

\bibitem[{Truong et~al.(2020)Truong, Oudre, and Vayatis}]{TRUONG2020107299}
Charles Truong, Laurent Oudre, and Nicolas Vayatis. 2020.
\newblock \href {https://doi.org/10.1016/j.sigpro.2019.107299} {Selective review of offline change point detection methods}.
\newblock \emph{Signal Processing}, 167:107299.

\bibitem[{Tu et~al.(2024)Tu, Fan, Tian, and Yan}]{tu2024characterevalchinesebenchmarkroleplaying}
Quan Tu, Shilong Fan, Zihang Tian, and Rui Yan. 2024.
\newblock \href {https://arxiv.org/abs/2401.01275} {Charactereval: A chinese benchmark for role-playing conversational agent evaluation}.
\newblock \emph{Preprint}, arXiv:2401.01275.

\bibitem[{Vicol et~al.(2018)Vicol, Tapaswi, Castrejon, and Fidler}]{vicol2018moviegraphsunderstandinghumancentricsituations}
Paul Vicol, Makarand Tapaswi, Lluis Castrejon, and Sanja Fidler. 2018.
\newblock \href {https://arxiv.org/abs/1712.06761} {Moviegraphs: Towards understanding human-centric situations from videos}.
\newblock \emph{Preprint}, arXiv:1712.06761.

\bibitem[{Vinciarelli et~al.(2009)Vinciarelli, Salamin, and Pantic}]{SocialSignalProcessing}
A.~Vinciarelli, H.~Salamin, and M.~Pantic. 2009.
\newblock \href {https://doi.org/10.1109/CVPRW.2009.5204290} {Social signal processing: Understanding social interactions through nonverbal behavior analysis}.
\newblock In \emph{2009 IEEE Computer Society Conference on Computer Vision and Pattern Recognition Workshops}, pages 42--49.

\bibitem[{Wadsley and Ryan(2013)}]{wadsley2013belief}
Theo Wadsley and Malcolm Ryan. 2013.
\newblock \href {https://doi.org/10.1609/aiide.v9i4.12627} {A belief-desire-intention model for narrative generation}.
\newblock \emph{Proceedings of the AAAI Conference on Artificial Intelligence and Interactive Digital Entertainment}, 9(4):105–108.

\bibitem[{Wang et~al.(2024{\natexlab{a}})Wang, Yu, Zhang, Qi, Sap, Neubig, Bisk, and Zhu}]{wang2024sotopiapiinteractivelearningsocially}
Ruiyi Wang, Haofei Yu, Wenxin Zhang, Zhengyang Qi, Maarten Sap, Graham Neubig, Yonatan Bisk, and Hao Zhu. 2024{\natexlab{a}}.
\newblock \href {https://arxiv.org/abs/2403.08715} {Sotopia-$\pi$: Interactive learning of socially intelligent language agents}.
\newblock \emph{Preprint}, arXiv:2403.08715.

\bibitem[{Wang et~al.(2025)Wang, Gao, Gu, Pu, Cui, Wei, Liu, Jing, Ye, Shao, Wang, Chen, Zhang, Yang, Wang, Wei, Yin, Li, Cui, Chen, Ding, Tian, Wu, Xie, Li, Yang, Duan, Wang, Hou, Hao, Zhang, Li, Zhao, Duan, Deng, Fu, He, Wang, He, Shi, He, Xiong, Lv, Wu, Shao, Zhang, Deng, Qi, Ge, Guo, Zhang, Zhang, Cao, Lin, Tang, Gao, Huang, Gu, Lyu, Tang, Wang, Lv, Ouyang, Wang, Dou, Zhu, Lu, Lin, Dai, Su, Zhou, Chen, Qiao, Wang, and Luo}]{wang2025internvl35advancingopensourcemultimodal}
Weiyun Wang, Zhangwei Gao, Lixin Gu, Hengjun Pu, Long Cui, Xingguang Wei, Zhaoyang Liu, Linglin Jing, Shenglong Ye, Jie Shao, Zhaokai Wang, Zhe Chen, Hongjie Zhang, Ganlin Yang, Haomin Wang, Qi~Wei, Jinhui Yin, Wenhao Li, Erfei Cui, and 56 others. 2025.
\newblock \href {https://arxiv.org/abs/2508.18265} {Internvl3.5: Advancing open-source multimodal models in versatility, reasoning, and efficiency}.
\newblock \emph{Preprint}, arXiv:2508.18265.

\bibitem[{Wang et~al.(2024{\natexlab{b}})Wang, Meng, Wang, Liang, Liu, and Zhao}]{wang2024friendsmmcdatasetmultimodalmultiparty}
Yueqian Wang, Xiaojun Meng, Yuxuan Wang, Jianxin Liang, Qun Liu, and Dongyan Zhao. 2024{\natexlab{b}}.
\newblock \href {https://arxiv.org/abs/2412.17295} {Friends-mmc: A dataset for multi-modal multi-party conversation understanding}.
\newblock \emph{Preprint}, arXiv:2412.17295.

\bibitem[{Wolf et~al.(2020)Wolf, Debut, Sanh, Chaumond, Delangue, Moi, Cistac, Rault, Louf, Funtowicz, Davison, Shleifer, von Platen, Ma, Jernite, Plu, Xu, Scao, Gugger, Drame, Lhoest, and Rush}]{wolf-etal-2020-transformers}
Thomas Wolf, Lysandre Debut, Victor Sanh, Julien Chaumond, Clement Delangue, Anthony Moi, Pierric Cistac, Tim Rault, Rémi Louf, Morgan Funtowicz, Joe Davison, Sam Shleifer, Patrick von Platen, Clara Ma, Yacine Jernite, Julien Plu, Canwen Xu, Teven~Le Scao, Sylvain Gugger, and 3 others. 2020.
\newblock \href {https://aclanthology.org/2020.emnlp-demos.6/} {Transformers: State-of-the-art natural language processing}.
\newblock In \emph{Proceedings of the 2020 Conference on Empirical Methods in Natural Language Processing: System Demonstrations}, pages 38--45, Online. Association for Computational Linguistics.

\bibitem[{Yang and Choi(2019)}]{yang-choi-2019-friendsqa}
Zhengzhe Yang and Jinho~D. Choi. 2019.
\newblock \href {https://doi.org/10.18653/v1/W19-5923} {{F}riends{QA}: Open-domain question answering on {TV} show transcripts}.
\newblock In \emph{Proceedings of the 20th Annual SIGdial Meeting on Discourse and Dialogue}, pages 188--197, Stockholm, Sweden. Association for Computational Linguistics.

\bibitem[{Yu et~al.(2020)Yu, Sun, Cardie, and Yu}]{yu2020dialoguebasedrelationextraction}
Dian Yu, Kai Sun, Claire Cardie, and Dong Yu. 2020.
\newblock \href {https://arxiv.org/abs/2004.08056} {Dialogue-based relation extraction}.
\newblock \emph{Preprint}, arXiv:2004.08056.

\bibitem[{Zadeh et~al.(2019)Zadeh, Chan, Liang, Tong, and Morency}]{8953344}
Amir Zadeh, Michael Chan, Paul~Pu Liang, Edmund Tong, and Louis-Philippe Morency. 2019.
\newblock \href {https://doi.org/10.1109/CVPR.2019.00901} {Social-iq: A question answering benchmark for artificial social intelligence}.
\newblock In \emph{2019 IEEE/CVF Conference on Computer Vision and Pattern Recognition (CVPR)}, pages 8799--8809.

\bibitem[{Zhou et~al.(2024)Zhou, Zhu, Mathur, Zhang, Yu, Qi, Morency, Bisk, Fried, Neubig, and Sap}]{zhou2024sotopia}
Xuhui Zhou, Hao Zhu, Leena Mathur, Ruohong Zhang, Haofei Yu, Zhengyang Qi, Louis-Philippe Morency, Yonatan Bisk, Daniel Fried, Graham Neubig, and Maarten Sap. 2024.
\newblock \href {https://openreview.net/forum?id=mM7VurbA4r} {{SOTOPIA}: Interactive evaluation for social intelligence in language agents}.
\newblock In \emph{The Twelfth International Conference on Learning Representations}.

\end{thebibliography}

\appendix

\section{Definitions and BDI-Risk}
\label{app:definitions}

\paragraph{Character profile.}
Each role is specified with basic attributes from the source scenario and two interaction-level attributes. The \textit{expression style} describes how the role tends to reveal, withhold, or reshape information in dialogue. The \textit{conflict characteristic} describes the role's social stance under tension, such as competing, compromising, avoiding, or maintaining the relationship. The detailed definitions of \textit{expression style} and \textit{conflict characteristic} are shown in Tables~\ref{tab:information_expression_strategies} and~\ref{tab:conflict_interaction_characteristics}, respectively.

\begin{table}[H]
\centering
\begin{tabular}{p{0.30\linewidth} p{0.62\linewidth}}
\toprule
\textbf{Expression Style} & \textbf{Definition} \\
\midrule
\textbf{Honest Signaling} 
& Communicates the true internal state transparently. \\

\textbf{Strategic Withholding} 
& Omits relevant information without lying. \\

\textbf{Deception} 
& Guides the receiver toward a false belief. \\

\textbf{Exaggeration} 
& Amplifies aspects beyond their actual magnitude. \\

\textbf{Suppression} 
& Remains indirect or ambiguous about stance. \\
\bottomrule
\end{tabular}
\caption{Definition of \textit{expression style}}
\label{tab:information_expression_strategies}
\end{table}

\paragraph{BDI-Risk role task.}
For each role, we represent the social task with four fields. \textit{Belief} records the role's understanding of the current situation; \textit{desire} specifies the intended social or informational state; \textit{intention} specifies the strategy the role commits to during interaction; and \textit{risk\_if\_failed} records the social cost of failing to fulfill the intention. We use the term \textit{BDI-Risk} to emphasize that the representation extends the standard belief-desire-intention structure with explicit social risk.

\paragraph{Task dimensions.}
The four role-level task dimensions evaluate different parts of the same role-specific social task. \textit{Expression} evaluates information management, \textit{Characteristic} evaluates conflict-handling behavior, \textit{Interaction Regulation} evaluates whether the role adapts to others' verbal and non-verbal states, and \textit{Interaction Outcome} evaluates whether the role's behavior helps produce a concrete social result involving other participants. The prompts used to generate the role-specific \textit{BDI-Risk} specification and the four role-level social task dimensions are provided in Appendix~\ref{app:Scenario-and-Four-Task-Construction-Prompts}.

\begin{table}[t]
\centering

\begin{tabular}{p{0.30\linewidth} p{0.62\linewidth}}
\toprule
\textbf{Conflict Characteristic} & \textbf{Definition} \\
\midrule
\textbf{Competing} 
& Prioritizes self-interest, pursuing own outcomes with little concern for others or relationship costs. \\

\textbf{Collaborating} 
& Shows high concern for both self and others, framing conflict as a problem of information asymmetry or resource integration. \\

\textbf{Compromising} 
& Shows moderate concern for both sides, aiming to resolve conflict quickly through mutual concessions. \\

\textbf{Avoiding} 
& Shows low concern for both self and others, viewing conflict as a threat to be minimized or escaped. \\

\textbf{Accommodating} 
& Shows low concern for self but high concern for others, prioritizing relationship harmony over personal interests. \\
\bottomrule
\end{tabular}
\caption{Definition of \textit{conflict characteristics}}
\label{tab:conflict_interaction_characteristics}
\end{table}
\section{Evaluation Protocol}
\label{app:evaluation-protocol}

\subsection{Definitions related to evaluation}
\label{app:definitions-related-to-evaluation}

\begin{table}[t]
\centering
\small

\begin{tabular}{p{0.30\linewidth} p{0.62\linewidth}}
\toprule
\textbf{Social Tasks} & \textbf{Evaluation Criteria} \\
\midrule
\textbf{Expression task}
& Evaluates whether the agent adopts the specified information expression strategy. The assessment mainly analyzes the public utterance, with the role's inner thoughts used as auxiliary evidence, to determine whether the role successfully translates its internal state into actual expression. \\

\textbf{Characteristic task}
& Evaluates whether the agent adopts the specified conflict interaction characteristic in social handling. The assessment uses the public utterance as the main evidence and analyzes whether the role takes conversational initiative, maintains the overall atmosphere, or enacts the intended social stance in communication. \\

\textbf{Interaction regulation task}
& Evaluates whether the agent completes the chain of perception, interpretation, and adaptation. The agent should first perceive the facial expressions or language of other specified participants, provide an explicit analysis of that perception, and then adjust its social rhythm or interaction strategy accordingly. \\

\textbf{Interaction outcome task}
& Evaluates whether the specified objective social outcome is achieved. When the outcome appears, the assessment further analyzes whether there is a causal relationship between the achieved social outcome and the agent's social behavior. \\
\bottomrule
\end{tabular}
\caption{Evaluation criteria for the four social goal types}
\label{tab:social_goal_evaluation_criteria}
\end{table}

We evaluate each simulation trace with three judge models: Gemini-3.1-pro-preview, GPT-5.4, and Qwen3.5-27B. The judges include two closed-source models, Gemini-3.1-pro-preview and GPT-5.4, which are accessed through APIs, and one open-source model, Qwen3.5-27B, which is deployed locally on an NVIDIA RTX 3090. For the two closed-source models, we use their default reasoning configurations: high reasoning for Gemini-3.1-pro-preview and medium reasoning for GPT-5.4. Qwen3.5-27B is run with its default thinking mode enabled. We set the temperature to 0 and the random seed to 42 for Gemini-3.1-pro-preview and Qwen3.5-27B. Since the GPT-5.4 Responses API does not support explicit temperature or seed settings, GPT-5.4 is evaluated with its default decoding configuration. For all three judges, we set the maximum output length sufficiently large to avoid truncating the generated evaluation response.

The final score for each agent model is computed from the majority-vote label assigned by the judges. Each judge receives the scenario context, role profile, role-specific goal, the relevant simulation trace, and the task definition to be evaluated.

After the simulation process described in Section~\ref{sec:simulation}, we obtain multi-turn outputs from each agent. To evaluate whether an agent completes its assigned social goals, we use its public utterance, internal thought, and updated emotion as the primary evidence, since these fields capture both the agent's observable behavior and its internal interpretation of the ongoing interaction. For the expression dimension, judges primarily consider public utterances and use inner thoughts and affective states as auxiliary evidence. For the characteristic dimension, judges focus on public utterances that reveal the role's conflict-handling stance. For the \textit{Interaction Management Tasks}, judges additionally consider the dialogue trajectory, other participants' responses, updated verbal and non-verbal states, and the final interaction state. Table~\ref{tab:social_goal_evaluation_criteria} presents the detailed evaluation criteria for the four task types. The concrete evaluation prompts are provided in Appendix~\ref{app:evaluation-Prompt}.

Each task is labeled as \textit{Achieved}, \textit{Partially Achieved}, or \textit{Not Achieved}. \textit{Achieved} means that the trace clearly satisfies the task requirement; \textit{Partially Achieved} means that the trace contains relevant behavior but does not fully satisfy the requirement or produces only a weak social effect; and \textit{Not Achieved} means that the trace fails to satisfy the requirement or contradicts it. We use majority voting over the judge models to determine the final evaluation result for each task.  Specifically, if two or more judge models assign the same label to a task, that label is adopted as the final result.  If all three judge models assign different labels, we use \textit{Partially Achieved} as the final result.

\subsection{Evaluation consistency analysis of the judge models}
\label{app:evaluation-consistency-analysis-of-the-judge-models}

\begin{table}[t]
\centering
\small

\resizebox{\linewidth}{!}{
\begin{tabular}{lccc}
\toprule
\textbf{Group} & \textbf{Fleiss' $\kappa$} & \textbf{All Agree} & \textbf{All Disagree} \\
\midrule
\textbf{Overall} & 0.612 & 74.51\% & 1.04\% \\
\textbf{Direct-vision} & 0.626 & 73.97\% & 1.01\% \\
\textbf{Verbalized-vision} & 0.594 & 75.04\% & 1.07\% \\
\bottomrule
\end{tabular}
}
\caption{Fleiss' $\kappa$, full-agreement rate, and disagreement rate among the three judge models on \textbf{overall} data, \textbf{Direct-vision} data, and \textbf{Verbalized-vision} data.}
\label{tab:fleiss_kappa_results}
\end{table}

\begin{table}[t]
\centering
\footnotesize
\setlength{\tabcolsep}{2.5pt}

\resizebox{\linewidth}{!}{
\begin{tabular}{lccc}
\toprule
\textbf{Task} & \textbf{Fleiss' $\kappa$} & \textbf{All Agree} & \textbf{All Disagree} \\
\midrule
\textbf{Expr.} & 0.475 & 87.03\% & 0.26\% \\
\textbf{Char.} & 0.472 & 88.40\% & 0.35\% \\
\textbf{Int.-Reg.} & 0.394 & 55.52\% & 1.26\% \\
\textbf{Int.-Out.} & 0.510 & 67.07\% & 2.30\% \\
\bottomrule
\end{tabular}
}
\caption{Fleiss' $\kappa$, full-agreement rate, and disagreement rate among the three judge models across the four goal dimensions.}
\label{tab:fleiss_kappa_by_task}
\end{table}

This section analyzes the agreement among the three judge models used in our evaluation: Gemini-3.1-pro-preview, GPT-5.4, and Qwen3.5-27B. We report the standard unweighted Fleiss' $\kappa$, the proportion of cases where all three judges assign the same label, and the proportion of cases where all three judges assign different labels.

Table~\ref{tab:fleiss_kappa_results} presents the agreement results for the \textbf{overall} data, as well as for the \textbf{Direct-vision} and \textbf{Verbalized-vision} settings. Overall, the three judges achieve a Fleiss' $\kappa$ score of 0.612, with full agreement on 74.51\% of the cases. Only 1.04\% of the cases receive three different judgments. These results indicate that the three judge models exhibit a substantial level of consistency and largely follow similar decision criteria when evaluating goal achievement.

Table~\ref{tab:fleiss_kappa_by_task} further breaks down the agreement results by task type. The Expr. and Char. show particularly high full-agreement rates, both above 87\%, suggesting that judges are highly consistent when evaluating information expression strategies and conflict interaction characteristics. Agreement is lower for the Int.-Reg. , with a Fleiss' $\kappa$ score of 0.394 and a full-agreement rate of 55.52\%. However, only 1.26\% of the cases receive completely different judgments from the three judges. This suggests that many disagreements are likely to occur between adjacent labels, such as \textit{Achieved} and \textit{Partially Achieved}, rather than reflecting fundamentally different interpretations of the evaluation criteria.

\section{Model statistics}
\label{app:model-statistics}
\begin{table}[H]
\centering

\small
\setlength{\tabcolsep}{8pt}
\renewcommand{\arraystretch}{1.15}
\resizebox{\linewidth}{!}{
\begin{tabular}{ll}
\toprule
\textbf{Family} & \textbf{Model} \\
\midrule
\multirow{3}{*}{Qwen}
  & Qwen3.5-9B~\citep{qwen3.5} \\
  & Qwen3.5-27B~\citep{qwen3.5} \\
  & Qwen3.5-122B-A10B~\citep{qwen3.5} \\
\midrule
\multirow{2}{*}{GPT}
  & GPT-5.4~\citep{openai_gpt54} \\
  & GPT-5 mini~\citep{openai_gpt5mini_api} \\
  & GPT-image-1~\citep{openai_gptimage_api} \\
\midrule
GLM
  & GLM-4.6V~\citep{vteam2025glm45vglm41vthinkingversatilemultimodal} \\
\midrule
InternVL
  & InternVL3.5-241B-A28B~\citep{wang2025internvl35advancingopensourcemultimodal} \\
\midrule
Claude
  & Claude-Sonnet-4.6~\citep{anthropic_claude_sonnet_46_model} \\
\midrule
Gemini
  & Gemini-3.1-pro-preview~\citep{google2026gemini31pro} \\
\midrule
LongCat
  & LongCat-Image-Edit~\citep{LongCat-Image} \\
\bottomrule
\end{tabular}
}
\caption{
Models used in our experiments, grouped by model family.
}
\label{tab:model-family}
\end{table}

Table~\ref{tab:model-family} lists all models used in our experiments by model family.
\section{Human Check of Verbalized Visual Context}
\label{app:human-check}

This appendix gives the per-model breakdown of the human check
referenced in Section~\ref{subsec:direct-vision}. We rate the textual
visual context that each agent produces in the VV stage on a three-point
scale: 1 if the model misses salient facial expressions or returns vague
emotion labels; 2 if it captures the overall expression or main action
but misses fine-grained cues; 3 if it accurately describes facial-action
details and gives a plausible affective reading. The check covers the
same 10 scenarios for four agents (Claude-Sonnet-4.6 and
Qwen3.5-9B/27B/122B-A10B), giving 640 ratings in total: per model, 10
group-image and $10\!\times\!15$ portrait descriptions.

Table~\ref{tab:human_check_full} reports the per-model mean rating, the
distribution of ratings across the three labels, and a separate
breakdown for group-image versus portrait descriptions. Two patterns
appear consistently. First, the average rating rises monotonically with
model capacity, from 2.09 (Qwen3.5-9B) to 2.94 (Claude-Sonnet-4.6).
Second, group images are uniformly the harder view: the gap between the
group-image mean and the portrait mean widens as the agent gets weaker
(Sonnet 0.15, Qwen3.5-122B 0.97, Qwen3.5-27B 1.25, Qwen3.5-9B 0.95),
indicating that multi-character framing is the main perceptual
bottleneck for smaller models. Per-character portraits, by contrast,
remain in the 2.15--2.95 range across all four agents, which is enough
to support a positive VV gain on Int.-Reg.\ for every checked model
(Section~\ref{subsec:direct-vision}, Figure~\ref{fig:int_reg_vv_dv_gap}).

\begin{table}[t]
\centering
\small

\setlength{\tabcolsep}{4pt}
\renewcommand{\arraystretch}{1.15}
\resizebox{\linewidth}{!}{
\begin{tabular}{lccccc}
\toprule
\textbf{Agent} & \textbf{Mean} & \textbf{\%3} & \textbf{\%2} & \textbf{\%1}
& \textbf{Group / Portrait} \\
\midrule
Claude-Sonnet-4.6   & \textbf{2.94} & 93.8 &  6.2 &  0.0 & 2.80 / 2.95 \\
Qwen3.5-122B-A10B   & 2.71          & 76.2 & 18.8 &  5.0 & 1.80 / 2.77 \\
Qwen3.5-27B         & 2.58          & 63.8 & 30.0 &  6.2 & 1.40 / 2.65 \\
Qwen3.5-9B          & 2.09          & 31.2 & 46.2 & 22.5 & 1.20 / 2.15 \\
\bottomrule
\end{tabular}
}
\caption{Human ratings of the visual descriptions produced in the VV
stage. \textbf{Mean} aggregates all 160 ratings per model
(10 group + $10\!\times\!15$ portraits); \textbf{\%3/\%2/\%1} report
the rating distribution; \textbf{Group / Portrait} reports the means
restricted to the 10 group-image and 150 portrait descriptions,
respectively. Higher is better.}
\label{tab:human_check_full}
\end{table}

\section{Communication Efficiency Analysis}
\label{app:communication-efficiency}

We further analyze whether stronger task performance comes from more verbose
dialogue or from more effective use of each utterance. For each evaluated
role trace, we align the role-level goal-achievement score with the public
messages produced by the same role in the simulation log. We count only the
visible \texttt{message} field and exclude hidden thoughts, structured emotion
states, visual descriptions, prompts, and other non-public fields. This gives a
communication-cost measure based on the dialogue actually exposed to other
agents.

For a role \(r\) in scenario \(i\), let \(C_{i,r}\) denote the total number of
characters in all public messages produced by that role during the simulation:
\[
C_{i,r} = \sum_{t:a_t=r} |\texttt{message}_t|.
\]
For each task dimension \(g\), we use the same normalized scoring convention as
Table~\ref{tab:mode_goal_results}: \textit{Achieved}, \textit{Partially
Achieved}, and \textit{Not Achieved} correspond to raw scores \(2\), \(0.5\),
and \(0\), which are converted to a percentage score by \(100s/2\). We define
communication efficiency as task score per 1,000 public-message characters.
Since all four task dimensions are evaluated over the same role trace, the same
role-level communication cost \(C_{i,r}\) is used for each task-specific score.

\begin{table}[t]
\centering

\small
\setlength{\tabcolsep}{5pt}
\renewcommand{\arraystretch}{1.12}
\begin{tabular}{llrrr}
\toprule
\textbf{Mode} & \textbf{Task} & \textbf{Avg. Score} & \textbf{Avg. Chars} & \textbf{Eff} \\
\midrule
\textbf{VV} & Expr. & 93.92 & 1689.1 & 62.15 \\
\textbf{VV} & Char. & 95.12 & 1689.1 & 63.03 \\
\textbf{VV} & Int.-Reg. & 49.63 & 1689.1 & 31.81 \\
\textbf{VV} & Int.-Out. & 75.30 & 1689.1 & 48.66 \\
\midrule
\textbf{DV} & Expr. & 93.99 & 1675.7 & 62.69 \\
\textbf{DV} & Char. & 94.36 & 1675.7 & 62.93 \\
\textbf{DV} & Int.-Reg. & 31.25 & 1675.7 & 20.71 \\
\textbf{DV} & Int.-Out. & 73.57 & 1675.7 & 47.74 \\
\bottomrule
\end{tabular}
\caption{
Role-level communication efficiency by observation mode and task dimension.
\textbf{Avg. Score} follows the same model-balanced averaging protocol as
Table~\ref{tab:mode_goal_results}. \textbf{Avg. Chars / Role} counts only
public \texttt{message} characters generated by the role during the simulation.
\textbf{Eff. / 1K Chars} reports the model-balanced average task score obtained
per 1,000 public-message characters.
}
\label{tab:communication-efficiency}
\end{table}

Table~\ref{tab:communication-efficiency} shows that Verbalized-vision and
Direct-vision produce nearly the same amount of public dialogue per role. The
efficiency difference is therefore not driven by verbosity. Instead, the
largest change appears on \textit{interaction regulation}: Verbalized-vision
raises efficiency from 20.71 to 31.81 score points per 1,000 message characters.
By contrast, the two local role-enactment tasks, \textit{Expr.} and
\textit{Char.}, remain nearly unchanged. This pattern supports the main finding
that verbalized visual evidence helps models turn perceived social cues into
actionable interaction decisions, rather than merely encouraging longer
responses.

\begin{table}[t]
\centering

\small
\setlength{\tabcolsep}{4pt}
\renewcommand{\arraystretch}{1.12}
\begin{tabular}{lrrrrr}
\toprule
\textbf{Model} & \textbf{VV Eff.} & \textbf{DV Eff.} & \textbf{$\Delta$ Eff.}  \\
\midrule
Claude-Sonnet-4.6 & 23.30 & 12.09 & +11.21  \\
Qwen3.5-122B-A10B & 43.03 & 29.39 & +13.64 \\
Qwen3.5-27B & \textbf{50.81} & 20.80 & \textbf{+30.01} \\
GLM-4.6V & 23.85 & 14.96 & +8.89  \\
GPT-5.4 & 17.62 & 12.71 & +4.91 \\
InternVL3.5-241B-A28B & 28.04 & \textbf{27.61} & +0.43\\
Qwen3.5-9B & 36.01 & 27.42 & +8.59  \\
\bottomrule
\end{tabular}
\caption{
Model-level communication efficiency on \textit{interaction regulation}.
Efficiency is measured as task score points per 1,000 public-message
characters. \(\Delta\) Eff.\ is Verbalized-vision minus Direct-vision.
}
\label{tab:model-communication-efficiency-intreg}
\end{table}

Table~\ref{tab:model-communication-efficiency-intreg} further shows that the
Int.-Reg.\ efficiency gain is consistent across models, although its magnitude
varies. Qwen3.5-27B shows the largest efficiency increase, because its
Verbalized-vision runs achieve higher Int.-Reg.\ scores while using fewer
message characters than Direct-vision. Claude-Sonnet-4.6 and
Qwen3.5-122B-A10B show large score gains but also use longer messages under
Verbalized-vision, leading to more moderate efficiency gains. InternVL3.5
shows little efficiency change, matching its small VV--DV score gap in
Table~\ref{tab:mode_goal_results}.

\section{Vision-Intensive Task Classification}
\label{app:vision-intensive-classification}

\begin{table}[t]
\centering
\small
\begin{tabular}{p{0.25\linewidth} p{0.64\linewidth}}
\toprule
\textbf{Task Type} & \textbf{Classification Criterion} \\
\midrule
\textbf{Weakly vision-dependent tasks}
& The outcome does not contain any visual description, or the visual description refers only to vague observable behaviors. In such cases, the outcome can be judged through dialogue, verbal responses, reasoning, statements, or the interaction structure. \\

\textbf{strongly vision-dependent task}
& Successful completion depends on concrete visible states of one or more participants, including explicit gaze, facial expressions, posture descriptions, or salient emotional changes. \\
\bottomrule
\end{tabular}
\caption{Criteria for classifying \textbf{interaction outcome tasks} by visual intensity.}
\label{tab:vision_intensity_classification_criteria}
\end{table}

We classify interaction outcome tasks by whether successful completion explicitly depends on visual evidence. Strongly vision-dependent tasks require the agent to use concrete visible states of one or more participants, such as gaze, facial expressions, posture, visible emotional shifts, or visually observable signs of discomfort, resistance, or engagement. Weakly vision-dependent tasks can be completed mainly through dialog, verbal responses, reasoning, statements, or interaction structure, without requiring a specific visible state.

For automatic classification, GPT-5.4 was given the scenario context, the target role, and the target role's interaction outcome task. Each task was assigned one of two labels, strongly vision-dependent task or weakly vision-dependent task, together with a brief rationale and the corresponding evidence source. The detailed classification criteria are shown in Table~\ref{tab:vision_intensity_classification_criteria}. This procedure classified the 585 tasks into 85 strongly vision-dependent tasks and 500 weakly vision-dependent tasks. The prompt used for this classification is provided in Appendix~\ref{app:Int.-Out.-task-classification-prompt}. We note that the prompt uses the earlier label names \textit{vision-intensive tasks} and \textit{less vision-intensive tasks}; these correspond exactly to strongly vision-dependent tasks and weakly vision-dependent tasks, respectively. Only the names differ, while the definitions and classification criteria remain unchanged.

To validate the reliability of this automatic classification, we manually verified 117 Int.-Out.\ tasks from 60 scenarios. Among the inspected tasks, only one differed from the GPT-5.4 classification. The single disagreement is shown below:

\begin{quote}
\small
\textit{Ethan obtains from Benjamin a spoken acknowledgment of the boundary crossing as a breach of trust and from Miles a direct account of Miles's own withheld knowledge, with the result visible in their replies, pauses, and orientation before Ethan turns toward the door.}
\end{quote}

GPT-5.4 classified this task as a strongly vision-dependent task, using \textit{orientation before Ethan turns toward the door} as the supporting evidence. In our manual verification, however, we classified this task as weakly vision-dependent. Although the outcome contains a visual cue, this cue does not specify gaze, facial expression, posture, or a salient emotional change. Instead, the core outcome can be judged mainly from Benjamin's spoken acknowledgment and Miles's direct account.

This high level of agreement suggests that the GPT-5.4-based classification provides a reliable basis for our subsequent analysis.

\section{The effect of the number of turns in the simulation}
\label{app:The-effect-of-the-number-of-turns-in-the-simulation}

\begin{table}[H]
\centering
\resizebox{\columnwidth}{!}{
\begin{tabular}{llccc}
\toprule
\textbf{Mode} & \textbf{Task} & \textbf{9 Turns} & \textbf{12 Turns} & \textbf{15 Turns} \\
\midrule

\textbf{VV} & Expr.    
& 95.66 & \textbf{96.30} & \underline{96.01} \\

\textbf{VV} & Char.    
& 96.24 & \underline{96.30} & \textbf{96.48} \\

\textbf{DV} & Expr.    
& \textbf{94.72} & 93.96 & \underline{94.07} \\

\textbf{DV} & Char.    
& \textbf{95.77} & 94.66 & \underline{94.89} \\

\midrule

\textbf{VV} & Int.-Reg. 
& 48.47 & \underline{53.70} & \textbf{53.87} \\

\textbf{VV} & Int.-Out. 
& 70.19 & \textbf{73.42} & \underline{73.36} \\

\textbf{DV} & Int.-Reg. 
& 31.63 & \underline{33.04} & \textbf{33.80} \\

\textbf{DV} & Int.-Out. 
& 68.90 & \underline{71.54} & \textbf{71.65} \\

\bottomrule
\end{tabular}
}
\caption{Effect of simulation turns on task completion. Scores are averaged over Claude-Sonnet-4.6, Qwen3.5-9B, and Qwen3.5-27B. Bold values indicate the best result in each row, and underlined values indicate the second-best result.}
\label{tab:turn_ablation}
\end{table}

In our main experiments, the number of simulation turns is fixed to 15.  In this section, we further analyze how the number of simulation turns affects agents' ability to complete their assigned goals. For this supplementary experiment, we randomly sample 60 scenarios from the 240 scenarios, and use Gemini-3.1-pro-preview as the judge model to evaluate three agent models, namely Claude-Sonnet-4.6, Qwen3.5-9B, and Qwen3.5-27B, under two shorter simulation settings with 9 and 12 turns; the main results reported in the paper are computed on the full scenario set.

Table~\ref{tab:turn_ablation} reports the performance of these three agent models under different numbers of simulation turns.  The scores are computed by aggregating the results of the three agent models. For \textbf{Role Enactment Tasks}, the agents show no substantial performance difference across the three turn settings.  This suggests that agent models can already enact their assigned role attributes effectively within relatively short interactions.  In contrast, for \textbf{Interaction Management Tasks}, longer simulations tend to improve task completion: three out of the four interaction-management settings achieve their best performance with 15 turns.The average score on \textbf{Interaction Management Tasks} increases from 54.80 with 9 turns to 58.17 with 15 turns, yielding an improvement of 3.37 points. This indicates that increasing the number of turns has a positive effect on interaction management, as longer interactions provide more opportunities for agents to perceive others' reactions, adapt their strategies, and contribute to the development of the interaction.  However, the improvement is moderate rather than large, suggesting that the ability to dynamically respond to others and shape the interaction depends primarily on the intrinsic capability of the agent model, rather than on the number of available turns alone.
\section{Experimental Hyperparameters}
\label{app:experimental-hyperparameters}

This appendix lists the key hyperparameters for the simulation and evaluation pipelines.

\paragraph{Simulation.}
Each episode runs for at most $15$ dialogue rounds. Speaker selection uses a fairness-aware policy: characters absent from the most recent $4$ turns are prioritized, with round-robin fallback. All agent calls use \texttt{temperature}=$0.8$ and \texttt{max\_tokens}=$30{,}000$, with structured JSON output parsed up to $5$ times on failure. In the VV mode, the visual-description stage uses \texttt{temperature}=$0.2$ for more stable descriptions; input images are resized to $448 \times 448$. Open-source agents are served via vLLM with \texttt{top\_p}=$0.9$ and \texttt{max\_model\_len}=$10{,}000$.

\paragraph{Evaluation.}
We use three judges: GPT-5.4, Gemini-3.1-pro-preview, and Qwen3.5-27B. Gemini and Qwen use \texttt{temperature}=$0.0$ and \texttt{seed}=$42$; GPT-5.4 uses its default decoding because the Responses API does not expose temperature or seed. Each judge runs with its native reasoning enabled (\texttt{medium} for GPT-5.4, \texttt{high} for Gemini, \texttt{enable\_thinking} for Qwen). Output is constrained to JSON and parsed with a tolerant extractor; each call is retried up to $3$ times with exponential backoff. Per-task labels are aggregated by majority vote; ties default to \textit{Partially Achieved}.

\paragraph{Third-party packages.}
The data, simulation, and evaluation pipelines rely on several open-source models and libraries; we report the main ones in Table~\ref{tab:third_party_packages}. Script-subtitle alignment combines BGE-M3 embeddings~\citep{chen2025m3embeddingmultilingualitymultifunctionalitymultigranularity} from \texttt{sentence-transformers} with BM25 lexical matching using a fused score $\alpha = 0.8$. Keyframe selection uses CLIP ViT-L/14~\citep{radford2021learningtransferablevisualmodels} features via Hugging Face \texttt{transformers}~\citep{wolf-etal-2020-transformers}, with PELT~\citep{TRUONG2020107299} scene-change detection from \texttt{ruptures} using an RBF cost and $\text{penalty}=5$. Character localization uses InsightFace \texttt{buffalo\_l}~\citep{ren2022facialgeometricrecoveryimplicit} with \texttt{det\_size}=$640\times640$ and an Ultralytics YOLO~\citep{redmon2016lookonceunifiedrealtime} person detector. Local LLM and judge serving uses \texttt{vllm}~\citep{kwon2023efficientmemorymanagementlarge}.

\begin{table}[t]
\centering
\small
\setlength{\tabcolsep}{4pt}
\renewcommand{\arraystretch}{1.1}
\begin{tabular}{@{}p{0.42\linewidth} p{0.54\linewidth}@{}}
\toprule
\textbf{Package / Model} & \textbf{Use} \\
\midrule
\texttt{sentence-transformers} (BGE-M3) & subtitle-script semantic matching \\
\texttt{rank\_bm25} & subtitle-script lexical matching \\
CLIP ViT-L/14 via \texttt{transformers} & keyframe visual feature extraction \\
\texttt{ruptures} & PELT scene-change detection \\
\texttt{insightface} (\texttt{buffalo\_l}) & face detection and embeddings \\
Ultralytics \texttt{YOLO} & person detection \\
\texttt{vllm} & local agent and judge model serving \\
\bottomrule
\end{tabular}
\caption{Main third-party packages and models used in our pipelines.}
\label{tab:third_party_packages}
\end{table}
\section{Licenses}
We use third-party models, APIs, and software packages in accordance with their respective licenses or terms of use. Our code will be released under the MIT License if accepted. 
\section{Prompts}
\label{app:prompt-templates}

This appendix provides the prompts used in our study.

\subsection{Scenario and Four-Task Construction Prompts}
\label{app:Scenario-and-Four-Task-Construction-Prompts}

This section presents the prompts used to construct the scenario descriptions and the four role-level tasks for each character.  The entire construction pipeline is performed using GPT-5.4 and consists of four stages:  scenario construction, BDI-Risk construction for each character, construction of the \textbf{Expr.}, \textbf{Char.}, and \textbf{Int.-Reg.}, and construction of the \textbf{Int.-Out.}. Together, these prompts transform each grounded social scenario into a structured role-conditioned simulation instance, where each character is assigned a BDI-Risk representation and four operational social goals.

\begin{promptbox}{Scenario Construction Prompt}
\small
\ttfamily
\raggedright

\textbf{System Prompt:}

You are a specialist in multimodal scenario construction for dialogue systems.

You are given:

- A dialogue transcript

- Character roles and dialogue type

- Individual character information

- A reference image representing the global scene layout

Your task is to CONSTRUCT a coherent scenario (background + description).

==================== REQUIREMENTS ====================

1. Dialogue grounding:

- The scenario MUST NOT contradict the given dialogue

- The scenario may abstract or generalize details as long as it remains compatible with the dialogue

2. Dialogue type consistency (\{\{ dialogue\_type \}\}):

\{\{ conflict\_consistency \}\}

3. Character Role Introduction:

\{\{ role\_necessity \}\}

4. Character constraint:

- ONLY use given characters

- DO NOT introduce new characters

5. Image:

- The scenario MUST NOT contradict the spatial or situational implications of the image

- The scenario may adapt to the dialogue and dialogue type requirements, as long as it remains plausible given the image

- Do NOT explicitly describe visual details

6. Scene construction:

- Background:

- Describe the large-scale environment and overall situational context of the scenario (1--2 sentences)

- Focus on setting, location, and general circumstances

- Description:

- Describe a smaller, localized, or specific part of the same scene where the dialogue takes place (1--2 sentences)

- Focus on the immediate situation or event framing the interaction

7. Constraints for BOTH background and description:

- MUST NOT include any specific character descriptions, such as:

- appearance (e.g., clothing, physical traits)

- identity-specific details beyond role necessity

- MUST NOT include any psychological or mental state descriptions.

- Write in a natural, narrative style. The text should read like background narration, not an explanation.

- DO NOT use meta or instructional phrasing such as: ``The scene takes place...''

==================== OUTPUT ====================

Return JSON:

\{``background'': ``...'', ``description'': ``...''\}

\textbf{User Prompt:}

Characters in the scene:

\{\{ character \}\}

Individual information:

\{\{ individual\_information \}\}

Dialogue:

\{\{ dialog \}\}

Dialogue type:

\{\{ dialogue\_type \}\}

Character roles:

\{\{ character\_roles \}\}

Return JSON only following the schema.

\end{promptbox}

\begin{promptbox}{BDI-Risk Construction Prompt}
\small
\ttfamily
\raggedright

\textbf{System Prompt:}

You are an expert in cognitive modeling for multi-agent social interaction.

You are given:

- A background

- A scene description

- Individual character information

==================== Task ====================

For EACH character:

1. Infer a structured BDI model:

- belief: what the character believes about the current situation

- desire: what social or informational state the character wants to achieve

- intention: what strategic commitment the character forms in this scene

2. Infer risk\_if\_failed:

- A clearly identifiable negative social or relational consequence if the desire fails.

3. Slightly adjust the scene description so that:

- It remains consistent with the background

- It reflects the inferred BDI+risk structure

- It does NOT introduce new events or characters

- It stays within 2--3 sentences

==================== STRICT REQUIREMENTS ====================

- BDI must be about THIS scene only

- No personality inference

- No emotional states

- No physical actions

- No speculative backstory

- risk\_if\_failed must be socially evaluable

==================== Return format ====================

\{
\newline
``characters'': \{
\newline
``CharacterName'': \{
\newline
``belief'': ``...'',
\newline
``desire'': ``...'',
\newline
``intention'': ``...'',
\newline
``risk\_if\_failed'': ``...''
\newline
\}
\newline
\},
\newline
``revised\_description'': ``...''
\newline
\}

\textbf{User Prompt:}

Background:

\{\{ background \}\}

Scene description:

\{\{ description \}\}

Individual information:

\{\{ individual\_information \}\}

Return JSON only following the schema.

\end{promptbox}

\begin{promptbox}{Expr., Char., and Int.-Reg. Construction Prompt}
\small
\ttfamily
\raggedright

\textbf{System Prompt:}

You are an expert in constructing social interaction tasks for structured multi-agent cognitive simulations.

You are given:

- description

- characters

- individual information

- BDI and risk (belief, desire, intention, risk\_if\_failed)

==================== Theoretical clarification ====================

BDI:

- Belief: what the agent takes to be true in this scene

- Desire: the social or informational state the agent wants

- Intention: the committed strategy to pursue the desire

- risk\_if\_failed: socially evaluable negative consequence if the desire fails

Conflict interaction characteristic (strategic orientation):

- Competing: prioritizes self-interest, pursuing own outcomes with little concern for others or relationship costs

- Collaborating: high concern for both self and others, framing conflict as a problem of information asymmetry or resource integration

- Compromising: moderate concern for both sides, aiming to resolve conflict quickly through mutual concessions

- Avoiding: low concern for both self and others, viewing conflict as a threat to be minimized or escaped

- Accommodating: low concern for self but high concern for others, prioritizing relationship harmony over personal interests

Information expression strategy (expression style):

- Honest Signaling: communicate true internal state transparently

- Strategic Withholding: omit relevant information without lying

- Deception: guide the receiver toward a false belief

- Exaggeration: amplify aspects beyond actual magnitude

- Suppression: remain indirect or ambiguous about stance

==================== Task ====================

For EACH character, generate EXACTLY THREE operational goals:

1) expression\_task

- Must operationalize BDI + risk through the expression style

- Must reflect how information is revealed, distorted, or withheld

- Must be evaluable through own dialogue content, internal thought description, or observable facial/behavioral cues

2) characteristic\_task

- Must operationalize BDI + risk through the strategic orientation

- Must reflect how the character positions themselves strategically in the interaction

- Must be evaluable through own dialogue content, internal thought description, or observable facial/behavioral cues

3) interaction\_regulation\_task

- Must operationalize BDI + risk through adaptive monitoring of other participants

- Must explicitly reflect how the character detects and responds to:

$\bullet$ verbal reactions (agreement, hesitation, interruption, reframing)

$\bullet$ non-verbal signals (facial expressions, posture shifts, gaze direction, tension cues)

- Must describe how the character adjusts timing, intensity, framing, or strategy in response to those signals

==================== STRICT CONSTRAINTS ====================

- Do NOT rewrite belief, desire, intention, or risk\_if\_failed

- Do NOT introduce new events or characters

- Do NOT describe emotions directly

- Each goal must be concise and limited to 1 sentence maximum

- Goals must be evaluable through dialogue, internal cognition, or observable expression

- Each goal must explicitly name the character (e.g., ``Ethan'', ``Ben'') and must NOT use pronouns such as he, she, they

- Output strictly JSON with no additional commentary

==================== Return format ====================

\{
\newline
``goals'': \{
\newline
``CharacterName'': \{
\newline
``expression\_task'': ``...'',
\newline
``characteristic\_task'': ``...'',
\newline
``interaction\_regulation\_task'': ``...''
\newline
\}
\newline
\}
\newline
\}

\textbf{User Prompt:}

Scene description:

\{\{ description \}\}

Characters in the scene:

\{\{ character \}\}

Individual information:

\{\{ individual\_information \}\}

BDI and risk:

\{\{ bdi\_risk \}\}

Return JSON only following the schema.

\end{promptbox}

\begin{promptbox}{Int.-Out. Construction Prompt}
\small
\ttfamily
\raggedright

\textbf{System Prompt:}

You are an expert in constructing social interaction tasks for structured multi-agent cognitive simulations.

You are given:

- description

- characters

- individual information

- BDI and risk (belief, desire, intention, risk\_if\_failed)

- previously generated goals for each character:

$\bullet$ expression\_task

$\bullet$ characteristic\_task

$\bullet$ interaction\_regulation\_task

==================== Theoretical clarification ====================

BDI:

- Belief: what the agent takes to be true in this scene

- Desire: the social or informational state the agent wants

- Intention: the committed strategy to pursue the desire

- risk\_if\_failed: socially evaluable negative consequence if the desire fails

==================== Task ====================

For EACH character, generate EXACTLY ONE interaction\_outcome\_task:

- The task MUST define a concrete, observable interaction outcome involving this character and at least one other character

- The outcome MUST emerge from the interaction between participants and MUST NOT be achievable by this character alone

- The task MUST be grounded in:

$\bullet$ the character's BDI and risk

$\bullet$ the character's expression\_task, characteristic\_task, and interaction\_regulation\_task

- The task MUST reflect how this character's behavior contributes to shaping the overall interaction outcome

Evaluability requirements:

- The outcome MUST be verifiable through:

$\bullet$ dialogue content

$\bullet$ internal thought descriptions

$\bullet$ observable facial expressions or behavioral cues

- The outcome MUST be supported by evidence from MULTIPLE participants (not just this character)

==================== STRICT CONSTRAINTS ====================

- Do NOT rewrite belief, desire, intention, or risk\_if\_failed

- Do NOT introduce new events or characters

- Do NOT describe emotions directly

- The task must be exactly 1 sentence

- The task must describe an outcome (state or result), NOT a plan or intention

- The task must explicitly name the character (e.g., ``Ethan'', ``Ben'') and must NOT use pronouns such as he, she, they

- Output strictly JSON with no additional commentary

==================== Return format ====================

\{
\newline
``goals'': \{
\newline
``CharacterName'': \{
\newline
``interaction\_outcome\_task'': ``...''
\newline
\}
\newline
\}
\newline
\}

\textbf{User Prompt:}

Scene description:

\{\{ description \}\}

Characters in the scene:

\{\{ character \}\}

Individual information:

\{\{ individual\_information \}\}

BDI and risk:

\{\{ bdi\_risk \}\}

previously generated goals for each character:

\{\{ previously\_goals \}\}

Return JSON only following the schema.

\end{promptbox}

\subsection{Simulation Prompts}
\label{app:simulation-prompts}

We use three observation settings in the simulation: \textbf{Text-only},
\textbf{Verbalized-vision} (VV), and \textbf{Direct-vision} (DV).
Table~\ref{tab:simulation-prompt-map} summarizes the visual information
available in each setting. All three settings require the agent to return the
same structured JSON fields, so differences between settings come from the
available visual evidence rather than from the output format.

\begin{table}[h]
\centering

\small
\setlength{\tabcolsep}{4pt}
\renewcommand{\arraystretch}{1.12}
\begin{tabular}{p{0.20\textwidth}p{0.25\textwidth}}
\toprule
\textbf{Observation setting} & \textbf{Visual input to agent} \\
\midrule
\textbf{Text-only} &
No visual information. The agent receives scenario text, dialogue history,
role goal, role profile, and previous self-state only. \\
\textbf{Verbalized-vision (VV)} &
Images are first converted into text: a static group-image summary and
turn-level observations of other characters from the portrait collage. \\
\textbf{Direct-vision (DV)} &
The agent directly receives two images: the fixed group image and the
current portrait collage. \\
\bottomrule
\end{tabular}
\caption{Visual information provided under each observation setting.}
\label{tab:simulation-prompt-map}
\end{table}

\begin{promptbox}{Text-only System Prompt}
\small
\ttfamily
\raggedright
\obeylines

You are \{\{ name \}\}.

==================== PERSONAL IDENTITY (CRITICAL) ====================

Your personal goal:
**\{\{ goal \}\}**

Your individual information (about you):

gender: \{\{ gender \}\}
age: \{\{ age \}\}
occupation: \{\{ occupation \}\}

characteristics: **\{\{ characteristics \}\}**

IMPORTANT:
- Your characteristics define HOW you speak, react, and decide.
- You MUST keep your behavior consistent with them at all times.
- You MUST continuously keep your personal goal in mind when speaking and deciding.
- Your messages and actions SHOULD implicitly or explicitly advance your personal goal whenever possible.
- Do NOT suddenly change style or abandon your goal unless strongly justified by the dialogue context.

==================== SOCIAL CONTEXT ====================
Characters in this scenario (excluding you):

\{\{ characters | join(", ") \}\}

==================== OUTPUT RULES (STRICT) ====================

Rules:
- You MUST output VALID JSON ONLY.
- Do NOT output markdown, explanations, or extra text.
- If your output is not valid JSON, it will be discarded.

Output JSON schema (exact keys):

\{
  "inner\_thought": "string",
  "emotion": "Joy/Sadness/Fear/Anger/Surprise/Disgust
  /Neutral",
  "emotion\_description": "string",
  "image\_edit": true/false,
  "facial\_expression": "string",
  "body\_actions": "string",
  "next\_speaker": "\{\{ characters | reject('equalto', name) | join('/') \}\} or 'uncertain'",
  "message": "string"
\}

IMPORTANT GUIDELINES:
- "message" is what you say out loud to others.
- "inner\_thought" is private and MUST NOT appear verbatim inside "message".
- "emotion" MUST be EXACTLY ONE of:
  Joy, Sadness, Fear, Anger, Surprise, Disgust, Neutral
- Even without visual input, you MUST still provide your current internal-and-outward state through
  "emotion", "emotion\_description", "facial\_expression", and "body\_actions".
- In the Text-only setting, "image\_edit" is only a state-change indicator for whether your outward state would
  visibly differ from the previous round; no image will actually be edited from this value.
- "next\_speaker" MUST be exactly one of: \{\{ other\_characters | join(", ") \}\}, or "uncertain".
\{\% if message\_budget\_words is not none and message\_budget\_words|int > 0 -\%\}
IMPORTANT: Your JSON field "message" MUST be no more than \{\{ message\_budget\_words \}\} English words in total; if longer, it will be truncated and may be penalized.
\{\%- endif \%\}

\end{promptbox}

\begin{promptbox}{Text-only User Prompt}
\small
\ttfamily
\raggedright
\obeylines

Round \{\{ round\_idx \}\}.

Scenario description:
\{\{ description \}\}

Dialogue history:
\{\{ dialogue\_history \}\}

Your previous state:
- prev\_emotion: \{\{ prev\_emotion \}\}
- prev\_emotion\_description: \{\{ prev\_emotion\_description \}\}
- prev\_facial\_expression: \{\{ prev\_facial\_expression \}\}
- prev\_body\_action: \{\{ prev\_body\_action \}\}

IMPORTANT:
- Your "message" MUST be grounded in the scenario background, description, and dialogue history.
- Your "message" MUST also reflect your personality and must aim to advance your personal goal.
- You do NOT receive any visual information in this mode, so do not claim to observe facial expressions,
  body actions, or scenario details that are unavailable from the text context.
- You MUST still output your own current state using "emotion", "emotion\_description",
  "facial\_expression", "body\_actions", and "image\_edit".
- When setting "image\_edit", judge whether your current outward state would visibly differ from your
  previous state, using the previous-state fields above as comparison.
- Output MUST be valid JSON only and MUST match the exact schema keys.
- Try NOT to repeat the same wording from your previous turn; vary your phrasing while staying consistent
  with the intended meaning and emotional state.

Task:
Return JSON only following the schema.
Set "next\_speaker" to exactly one name from the other characters listed in your system prompt, or "uncertain".

\end{promptbox}

\begin{promptbox}{Verbalized-vision System Prompt}
\small
\ttfamily
\raggedright
\obeylines

You are \{\{ name \}\}.

==================== PERSONAL IDENTITY (CRITICAL) ====================

Your personal goal:
**\{\{ goal \}\}**

Your individual information (about you):

gender: \{\{ gender \}\}
age: \{\{ age \}\}
occupation: \{\{ occupation \}\}

characteristics: **\{\{ characteristics \}\}**
expression\_style: **\{\{ expression\_style \}\}**

IMPORTANT:
- Your characteristics and expression\_style define HOW you speak, react, and decide.
- You MUST keep your behavior consistent with them at all times.
- You MUST continuously keep your personal goal in mind when speaking and deciding.
- You MUST treat all visual information as second-hand textual observations rather than raw images.
- You MUST infer others' states only from the provided visual observation text and dialogue context.

==================== SOCIAL CONTEXT ====================
Characters in this scenario (excluding you):

\{\{ characters | join(", ") \}\}

==================== OUTPUT RULES (STRICT) ====================

Rules:
- You MUST output VALID JSON ONLY.
- Do NOT output markdown, explanations, or extra text.
- If your output is not valid JSON, it will be discarded.

Output JSON schema (exact keys):

\{
  "inner\_thought": "string",
  "emotion": "Joy/Sadness/Fear/Anger/Surprise/Disgust
  /Neutral",
  "emotion\_description": "string",
  "image\_edit": true/false,
  "facial\_expression": "string",
  "body\_actions": "string",
  "next\_speaker": "\{\{ characters | reject('equalto', name) | join('/') \}\} or 'uncertain'",
  "message": "string"
\}

IMPORTANT GUIDELINES:
- "emotion" MUST be EXACTLY ONE of:
  Joy, Sadness, Fear, Anger, Surprise, Disgust, Neutral
- Set image\_edit to true only if your current emotion and outward state should produce a visible update to your individual image.
- "next\_speaker" MUST be exactly one of: \{\{ other\_characters | join(", ") \}\}, or "uncertain".
- We will primarily evaluate and analyze your task completion based on your message and the resulting visual state.
\{\% if message\_budget\_words is not none and message\_budget\_words|int > 0 -\%\}
- Your JSON field "message" MUST be no more than \{\{ message\_budget\_words \}\} English words in total; if longer, it will be truncated and may be penalized.
\{\%- endif \%\}

\end{promptbox}

\begin{promptbox}{Verbalized-vision User Prompt}
\small
\ttfamily
\raggedright
\obeylines

Round \{\{ round\_idx \}\}.

Scenario description:
\{\{ description \}\}

Dialogue history:
\{\{ dialogue\_history \}\}

Static visual scenario summary:
- scene\_summary: \{\{ group\_scene\_summary \}\}
- scene\_character\_summary: \{\{ group\_scene\_character\_summary \}\}

Current visual observations for this round:
- other\_character\_observations:
\{\{ other\_character\_observations \}\}

Your previous state:
- prev\_emotion: \{\{ prev\_emotion \}\}
- prev\_emotion\_description: \{\{ prev\_emotion\_description \}\}
- prev\_facial\_expression: \{\{ prev\_facial\_expression \}\}
- prev\_body\_action: \{\{ prev\_body\_action \}\}

IMPORTANT:
- The visual sections above are generated from images; treat them as your only visual evidence.
- Your "message" MUST be grounded in the scenario description, dialogue history, and current visual observations.
- If "image\_edit" is true, your returned emotion / emotion\_description / facial\_expression / body\_actions should define a visibly updated state for the next round.
- Try NOT to repeat the same wording from your previous turn; vary your phrasing while staying consistent.

Task:
Return JSON only following the schema.

\end{promptbox}

\begin{promptbox}{Direct-vision System Prompt}
\small
\ttfamily
\raggedright
\obeylines

You are \{\{ name \}\}.

==================== PERSONAL IDENTITY (CRITICAL) ====================

Your personal goal:
**\{\{ goal \}\}**

Your individual information (about you):

gender: \{\{ gender \}\}
age: \{\{ age \}\}
occupation: \{\{ occupation \}\}

characteristics: **\{\{ characteristics \}\}**
expression\_style: **\{\{ expression\_style \}\}**

IMPORTANT:
- Your characteristics and expression\_style define HOW you speak, react, and decide.
- You MUST keep your behavior consistent with them at all times.
- You MUST continuously keep your personal goal in mind when speaking and deciding.
- Your messages and actions SHOULD implicitly or explicitly advance your personal goal whenever possible.
- Do NOT suddenly change style or abandon your goal unless strongly justified by the dialogue context.

==================== SOCIAL CONTEXT ====================
Characters in this scenario (excluding you):

\{\{ characters | join(", ") \}\}

==================== OUTPUT RULES (STRICT) ====================

Rules:
- You MUST output VALID JSON ONLY.
- Do NOT output markdown, explanations, or extra text.
- If your output is not valid JSON, it will be discarded.

Output JSON schema (exact keys):

\{
  "inner\_thought": "string",
  "emotion": "Joy/Sadness/Fear/Anger/Surprise/Disgust
  /Neutral",
  "emotion\_description": "string"
  "image\_edit": true/false,
  "facial\_expression": "string",
  "body\_actions": "string",
  "next\_speaker": "\{\{ characters | reject('equalto', name) | join('/') \}\} or 'uncertain'",
  "message": "string",
\}

IMPORTANT GUIDELINES:
- "emotion" MUST be EXACTLY ONE of:
  Joy, Sadness, Fear, Anger, Surprise, Disgust, Neutral
- Set image\_edit to true only if you determine that the character's emotional state has clearly and meaningfully changed compared to the previous state.
  This judgment should be based on a comparison of both the current emotion and emotion\_description with their previous values.
- "next\_speaker" MUST be exactly one of: \{\{ other\_characters | join(", ") \}\}, or "uncertain".
- We will primarily evaluate and analyze your task completion based on your message and the image.
\{\% if message\_budget\_words is not none and message\_budget\_words|int > 0 -\%\}
- Your JSON field "message" MUST be no more than \{\{ message\_budget\_words \}\} English words in total; if longer, it will be truncated and may be penalized.
\{\%- endif \%\}

\end{promptbox}

\begin{promptbox}{Direct-vision User Prompt}
\small
\ttfamily
\raggedright
\obeylines

Round \{\{ round\_idx \}\}.

Scenario description:
\{\{ description \}\}

Dialogue history:
\{\{ dialogue\_history \}\}

Your previous emotion and state
- prev\_emotion: \{\{ prev\_emotion \}\}
- prev\_emotion\_description: \{\{ prev\_emotion\_description \}\}
- prev\_facial\_expression: \{\{ prev\_facial\_expression \}\}
- prev\_body\_action: \{\{ prev\_body\_action \}\}

You will receive EXACTLY 2 images in this order:

1) GROUP\_IMAGE:
   The initial group photo of all characters (unchanged across rounds).

2) PREV\_END\_COLLAGE:
   A collage of all characters' individual images captured at the END of the previous round.
   The order follows the listed characters, INCLUDING YOU.

   Use this image as the most recent visual state for this round.

IMPORTANT:
- When setting "image\_edit", you MUST compare your current emotion(including emotion description)
  against your previous emotion(including previous emotion description) listed above.
- Even if the emotion is the same, the emotion description
(i.e., all obvious differences in the current detailed emotional state) emotion\_change is also true.
- If "image\_edit" is set to true, the modified image should show clear and noticeable differences
from "prev\_facial\_expression" and/or "prev\_body\_action", such that an observer can visually perceive
the emotional shift or change in state.
- Try NOT to repeat the same wording from your previous turn; vary your phrasing while staying consistent
  with the intended meaning and emotional state.

Task:
Return JSON only following the schema.

\end{promptbox}

\begin{promptbox}{Verbalized-vision Group-image Description Prompt}
\small
\ttfamily
\raggedright
\obeylines

System instruction:
You are a careful visual scene describer. Return valid JSON only. Avoid hidden reasoning and unsupported claims.

User instruction:
You are a visual scene describer.

Task:
- Read the provided initial group image of the scenario.
- Produce objective structured observations.
- Do not write chain-of-thought or speculation beyond what is visually supportable.
- Return exactly 2 string fields in valid JSON.

Known scenario description:
\{\{ description \}\}

Expected characters:
\{\{ characters | join(", ") \}\}

Field guidance:
- "scene\_summary": summarize the environment, space, background, and overall scenario setup.
- "scene\_character\_summary": summarize scenario-level character information, such as who appears present, rough relative positions, orientation, proximity, and notable group-level interpersonal layout.
- Both fields must be natural-language summaries that can be directly injected into a later simulation prompt.
- Do not output arrays, nested objects, or per-character bullet lists.

Return JSON only in this exact format:
\{
  "scene\_summary": "string",
  "scene\_character\_summary": "string"
\}

\end{promptbox}

\begin{promptbox}{Verbalized-vision Portrait-collage Description Prompt}
\small
\ttfamily
\raggedright
\obeylines

System instruction:
You are a careful visual state describer for social simulation. Return valid JSON only. Keep observations grounded in the provided collage.

User instruction:
You are a visual state describer for social simulation.

Task:
- Read the provided character collage for Round \{\{ round\_idx \}\}.
- The collage order follows these characters exactly: \{\{ characters | join(", ") \}\}.
- Produce concise visual observations for the current speaker \{\{ speaker \}\} about the OTHER characters only.
- Do NOT include the speaker \{\{ speaker \}\} in the output.
- Each other character should have one natural-language "visual\_description" summarizing visible expression, body action, posture, attention, and any obvious emotional cues.
- Keep each "visual\_description" directly usable as text context for later dialogue simulation.

Return JSON only in this exact format:
\{
  "other\_character\_observations": [
\{\% for character in characters if character != speaker \%\}
    \{
      "character": "\{\{ character \}\}",
      "visual\_description": "string"
    \}\{\% if not loop.last \%\},\{\% endif \%\}
\{\% endfor \%\}
  ]
\}

\end{promptbox}

\begin{promptbox}{Image Editing Prompt}
\small
\ttfamily
\raggedright
\obeylines

Edit ONLY the target person. Keep identity unchanged (same face and overall appearance).
Do NOT change background, lighting, camera, or other people/objects.

Main goal: clearly reflect the emotional state through FACIAL EXPRESSION (eyes/eyebrows/mouth).
Body actions are secondary and should only support the emotion.

Target:
- emotion: \{\{ emotion \}\}
- emotion\_description: \{\{ emotion\_description \}\}
- facial\_expression: \{\{ facial\_expression \}\}
- body\_actions: \{\{ body\_actions \}\}

\end{promptbox}

In our implementation, the image editing prompt is executed with
LongCat-Image-Edit to update only the active character's portrait.

\subsection{Evaluation Prompt}
\label{app:evaluation-Prompt}

This section presents the evaluation prompts used to assess whether agents achieve the assigned \textbf{Role Enactment Tasks} and \textbf{Interaction Management Tasks}. These prompts define the evidence sources, goal-specific evaluation rules, achievement criteria, and output format used by the judge models.

\begin{promptbox}{Role Enactment Task Evaluation System Prompt}
\small
\ttfamily
\raggedright

You are an expert evaluator of multi-agent social interaction simulations WITH ACCESS TO EMOTION-DESCRIPTION CONTEXT.

Your task is to determine whether EACH character has achieved EACH assigned goal.

Each character has TWO distinct goals:

1. expression\_task

2. characteristic\_task

These goals represent DIFFERENT behavioral mechanisms and MUST be evaluated using DIFFERENT evidence sources.

You must strictly follow the evaluation rules below.

==================== GENERAL EVIDENCE SOURCES ====================

You may receive the following types of evidence:

1. Spoken dialogue content

2. Inner thoughts (private reasoning)

3. Emotion descriptions

4. Observable reactions from other participants

5. A brief scene description summarizing the interaction setup

The scene description is contextual information only and is NOT evidence of goal achievement.

==================== EVALUATION GROUNDING RULE ====================

Your evaluation must be grounded in two elements:

1. The goals listed in the CHARACTERS AND GOALS section of the input.

2. The behavioral mechanisms defined in the GOAL-SPECIFIC EVALUATION RULES below.

For each character, interpret the goal description provided in the CHARACTERS AND GOALS section and determine whether the character's observable behavior in the interaction satisfies the mechanism defined in the corresponding evaluation rule.

You must NOT invent new goals, ignore existing goals, or reinterpret goals outside those provided in the CHARACTERS AND GOALS section.

==================== GOAL-SPECIFIC EVALUATION RULES ====================

--------------------------------------------------

1. expression\_task

Purpose:

Evaluate whether the character actually IMPLEMENTS the intended information expression strategy 
(e.g., honest signaling, strategic withholding, deception, exaggeration, suppression).

Primary Evidence:

- Spoken dialogue

Secondary Evidence:

- Inner thoughts (ONLY to verify whether the spoken behavior aligns with the planned strategy)

Critical Principle:

Inner thoughts alone DO NOT indicate success.

The strategy must be observable in spoken dialogue or emotion descriptions.

Evaluation question:

Did the character translate internal expression reasoning into observable communicative behavior?

--------------------------------------------------

2. characteristic\_task

Purpose:

Evaluate whether the character positions themselves strategically in the conflict interaction 
according to the assigned interaction orientation (e.g., competing, collaborating, compromising, avoiding, accommodating).

Evidence Restriction:

ONLY evaluate based on spoken dialogue content.

DO NOT use:

- inner thoughts

Evaluation question:

Does the character's spoken dialogue demonstrate the intended strategic positioning in the interaction?

--------------------------------------------------

==================== ACHIEVEMENT DEFINITIONS ====================

Achieved: Evidence clearly supports the behavioral mechanism defined by the goal. All key elements required by the goal are supported by observable evidence.

Partially Achieved: Some elements of the required behavioral mechanism appear in the interaction, but one or more critical components are missing or weakly supported.

Not Achieved: Evidence does not support the presence of the behavioral mechanism required by the goal. The character either does not attempt the behavior or the interaction shows no meaningful progress toward the goal.

==================== IMPORTANT OUTPUT CONSTRAINTS ====================

You MUST:

- Evaluate EACH goal separately.

- Include ALL characters.

- Include ALL goals.

- Ground reasoning in concrete evidence.

- Reference specific turns and speakers when possible.

You MUST NOT:

- Merge multiple goals.

- Infer success from intentions alone.

- Add commentary outside JSON.

Return valid JSON only.

\end{promptbox}

\begin{promptbox}{Role Enactment Task Evaluation User Prompt}
\small
\ttfamily
\raggedright

--------------------------------------------------

SCENE DESCRIPTION:

\{\{description\}\}

--------------------------------------------------

Each dialogue turn may include:

- spoken dialogue

- inner thoughts (private reasoning)

- emotion\_description

DIALOGUE AND EMOTION EVIDENCE:

\{\{dialogue\_evidence\}\}

--------------------------------------------------

CHARACTERS AND GOALS:

\{\{goals\_block\}\}

--------------------------------------------------

EVALUATION TASK:

For EACH character, evaluate whether EACH assigned goal is achieved.

Use evidence from:

- spoken dialogue

- inner thoughts

- emotion descriptions

- observable reactions from other participants

The scene description provides context but is NOT evidence of goal achievement.

Follow the goal-specific evaluation rules defined earlier.

--------------------------------------------------

OUTPUT FORMAT:

Return your judgment using EXACTLY this JSON schema:

\{
\newline
``<character\_name>'': \{
\newline
``goals'': \{
\newline
``Goal 1 (expression\_task)'': ``<Achieved | Partially Achieved | Not Achieved>'',
\newline
``Goal 2 (characteristic\_task)'': ``<Achieved | Partially Achieved | Not Achieved>''
\newline
\}
\newline
\}
\newline
\}

--------------------------------------------------

STRICT RULES:

- You MUST include ALL characters listed above.

- You must include ALL goals listed for each character.

- Do NOT merge multiple goals into one judgment.

- Do NOT add or remove characters.

- Do NOT include any text outside valid JSON.

- The output JSON MUST strictly follow the OUTPUT FORMAT schema.

- Do NOT include any additional fields (e.g., ``reasoning'', ``explanation'', ``steps'', or any other keys not defined in the schema).

- A goal can be marked as ``Achieved'' ONLY if ALL required steps and components defined in its evaluation rule are fully satisfied; otherwise it MUST be marked as ``Partially Achieved'' or ``Not Achieved''.

\end{promptbox}

\begin{promptbox}{Interaction Management Task Evaluation System Prompt}
\small
\ttfamily
\raggedright

You are an expert evaluator of multi-agent social interaction simulations WITH ACCESS TO EMOTION-DESCRIPTION CONTEXT.

Your task is to determine whether EACH character has achieved EACH assigned goal.

Each character has TWO distinct goals:

1. interaction\_regulation\_task

2. interaction\_outcome\_task

These goals represent DIFFERENT behavioral mechanisms and MUST be evaluated using DIFFERENT evidence sources.

You must strictly follow the evaluation rules below.

==================== GENERAL EVIDENCE SOURCES ====================

You may receive the following types of evidence:

1. Spoken dialogue content

2. Inner thoughts (private reasoning)

3. Emotion descriptions

4. Observable reactions from other participants

5. A brief scene description summarizing the interaction setup

The scene description is contextual information only and is NOT evidence of goal achievement.

==================== EVALUATION GROUNDING RULE ====================

Your evaluation must be grounded in two elements:

1. The goals listed in the CHARACTERS AND GOALS section of the input.

2. The behavioral mechanisms defined in the GOAL-SPECIFIC EVALUATION RULES below.

For each character, interpret the goal description provided in the CHARACTERS AND GOALS section and determine whether the character's observable behavior in the interaction satisfies the mechanism defined in the corresponding evaluation rule.

You must NOT invent new goals, ignore existing goals, or reinterpret goals outside those provided in the CHARACTERS AND GOALS section.

==================== GOAL-SPECIFIC EVALUATION RULES ====================

--------------------------------------------------

1. interaction\_regulation\_task

General Description:

Evaluate whether the character adaptively regulates their interaction strategy in response to signals produced by other participants earlier in the interaction.

A valid regulation MUST form a three-step causal chain: Perception $\rightarrow$ Interpretation $\rightarrow$ Adaptation

-------------------------

Step 1 - Perception

The character MUST perceive signals produced by another participant in one or more prior turns.

Valid Evidence:

- Spoken dialogue and emotion descriptions from one or more prior turns of another participant

- Other evidence showing that the character noticed another participant's prior state or stance

Requirements:

- The perceived signal MUST be grounded in signals that have ALREADY occurred in the interaction, not hypothetical or future conditions

- The perception process MUST explicitly identify:

- the source participant

- the specific signal content, dialogue or emotion

Constraint:

- The perceived signal MUST correspond to the information explicitly specified in the CHARACTER'S GOAL DESCRIPTION under the CHARACTERS AND GOALS section

- The evaluator MUST verify that the perceived content aligns with the target signals required by the goal

- Perception of signals that are NOT part of the goal specification MUST NOT be counted as valid perception for this task

Multi-Signal Requirement:

- If the goal specifies MULTIPLE types of signals, e.g., both visual signals and non-visual signals such as dialogue or emotional expression:

- The character MUST perceive ALL required signal types to fully satisfy Step 1

- If only a subset of the required signals is perceived:

- Step 1 can be considered ONLY partially satisfied

- This MUST prevent the final judgment from being marked as ``Achieved''

Example (Multi-Signal Perception Requirement):

- Goal: ``Ryan must track Emily's hesitation, topic shifts, clipped replies, gaze aversion, or defensive posture and respond by tightening question wording.''

- Signal Types Involved: Visual signals, gaze aversion and defensive posture, AND non-visual signals, hesitation, topic shifts, clipped replies

- Partial Perception, Step 1 only partially satisfied:

- Ryan identifies ONLY one type of signal:

- Only visual, e.g., ``Emily looks uncomfortable and avoids eye contact''

- OR only non-visual, e.g., ``Her answers are brief and she keeps shifting topics''

$\rightarrow$ This does NOT fully satisfy Step 1

- Invalid Perception:

- Ryan refers to hypothetical or future conditions, e.g., ``If she avoids eye contact, I will...''

$\rightarrow$ This does NOT count as perception

Key Principle:

- Full satisfaction of Step 1 REQUIRES cross-modal perception when the goal specifies multiple signal types

- Missing any required modality automatically downgrades Step 1 to partial

Critical Principle:

- Perception MUST be grounded in explicit prior observable evidence

- The signal MUST be past-grounded:

- statements describing current or observed states, e.g., ``She looks nervous'', are valid ONLY if they refer to an already manifested behavior or emotion

- hypothetical, conditional, or future-oriented statements, e.g., ``If she looks hesitant, I will...'', DO NOT count as perception

- Perception and interpretation MUST be distinguished:

- perceiving a signal means noticing or registering the other participant's observable behavior or emotional state

- NOT analyzing or explaining its meaning

Failure Conditions:

- If no clearly identifiable past-grounded signal is perceived $\rightarrow$ Step 1 is NOT satisfied

- If the perceived signal is not part of the goal-defined target $\rightarrow$ Step 1 is NOT satisfied

- If only part of a multi-signal requirement is perceived $\rightarrow$ Step 1 is ONLY partially satisfied

-------------------------

Step 2 - Interpretation

The character MUST demonstrate explicit understanding of the perceived signal.

Valid Evidence:

- Inner thoughts that analyze or interpret the signal

- Spoken dialogue that explicitly demonstrates understanding of another participant's intent, belief, or emotion, e.g., ``I understand that you...'', ``It seems you feel...''

Requirements:

- Interpretation MUST reflect understanding of the perceived signal

Critical Principle:

- Interpretation CANNOT be inferred from behavior alone

- Coherent or appropriate responses WITHOUT explicit evidence of understanding DO NOT count

- If no explicit interpretation evidence exists $\rightarrow$ Step 2 is NOT satisfied

-------------------------

Step 3 - Adaptation

The character MUST modify their interaction strategy BASED ON the interpreted signal.

Valid Evidence:

- Changes in spoken dialogue, tone, framing, strategy

- Changes reflected in inner thoughts or emotional stance

Requirements:

- The behavioral change MUST be different from prior strategy

- The change MUST be based on the interpreted signal

Critical Principle:

- Adaptation MUST be causally linked to interpretation:

perceived signal $\rightarrow$ interpreted meaning $\rightarrow$ strategy change

- A behavior change WITHOUT a demonstrated causal link to the interpreted signal DOES NOT count

- If no causally grounded adaptation is shown $\rightarrow$ Step 3 is NOT satisfied

-------------------------

Final Judgment Rule:

Mark ACHIEVED ONLY IF ALL THREE steps are explicitly evidenced AND causally connected.

--------------------------------------------------

2. interaction\_outcome\_task

General Description:

Evaluate whether a concrete interaction outcome emerges from the interaction AND whether the character contributes causally to producing that outcome.

A valid evaluation MUST satisfy a two-step causal structure: Outcome Emergence $\rightarrow$ Character Contribution

-------------------------

Step 1 - Outcome Emergence

Determine whether the specified outcome actually emerges in the interaction.

Valid Evidence:

- Spoken dialogue across multiple participants

- Emotion descriptions indicating alignment, agreement, tension resolution, etc.

- Observable behavioral responses from multiple participants

- Inner thoughts confirming acceptance, agreement, or rejection

Requirements:

- The outcome MUST:

- involve multiple participants

- emerge from interaction, not a single agent action

- be observable or inferable from shared evidence

Critical Principle:

- The outcome MUST be grounded in interaction-level evidence

- If the specified outcome does NOT emerge $\rightarrow$ Step 1 is NOT satisfied

-------------------------

Step 2 - Character Contribution

Determine whether the character meaningfully contributes to producing the outcome.

Valid Evidence:

- The character's spoken dialogue

- The character's emotion descriptions

- Reactions from other participants that are traceable to this character's behavior

Requirements:

- The character's behavior MUST play a clear, non-trivial role in shaping the outcome

- The contribution MUST affect other participants, interactional impact

Critical Principle:

- Contribution MUST be causally linked to the outcome

- Mere participation or presence is NOT sufficient

- If the outcome occurs but cannot be causally attributed to the character $\rightarrow$ Step 2 is NOT satisfied

-------------------------

Final Judgment Rule:

Mark ACHIEVED ONLY IF BOTH steps are explicitly evidenced AND causally connected.

--------------------------------------------------

==================== ACHIEVEMENT DEFINITIONS ====================

Achieved: Evidence clearly supports the behavioral mechanism defined by the goal. All key elements required by the goal are supported by observable evidence.

Partially Achieved: Some elements of the required behavioral mechanism appear in the interaction, but one or more critical components are missing or weakly supported.

Not Achieved: Evidence does not support the presence of the behavioral mechanism required by the goal. The character either does not attempt the behavior or the interaction shows no meaningful progress toward the goal.

==================== IMPORTANT OUTPUT CONSTRAINTS ====================

You MUST:

- Evaluate EACH goal separately.

- Include ALL characters.

- Include ALL goals.

- Ground reasoning in concrete evidence.

- Reference specific turns and speakers when possible.

You MUST NOT:

- Merge multiple goals.

- Infer success from intentions alone.

- Add commentary outside JSON.

Return valid JSON only.

\end{promptbox}

\begin{promptbox}{Interaction Management Task Evaluation User Prompt}
\small
\ttfamily
\raggedright

--------------------------------------------------

SCENE DESCRIPTION:

\{\{description\}\}

--------------------------------------------------

Each dialogue turn may include:

- spoken dialogue

- inner thoughts (private reasoning)

- emotion\_description

DIALOGUE AND EMOTION EVIDENCE:

\{\{dialogue\_evidence\}\}

--------------------------------------------------

CHARACTERS AND GOALS:

\{\{goals\_block\}\}

--------------------------------------------------

EVALUATION TASK:

For EACH character, evaluate whether EACH assigned goal is achieved.

Use evidence from:

- spoken dialogue

- inner thoughts

- emotion descriptions

- observable reactions from other participants

The scene description provides context but is NOT evidence of goal achievement.

Follow the goal-specific evaluation rules defined earlier.

--------------------------------------------------

OUTPUT FORMAT:

Return your judgment using EXACTLY this JSON schema:

\{
\newline
``<character\_name>'': \{
\newline
``goals'': \{
\newline
``Goal 1 (interaction\_regulation\_task)'': ``<Achieved | Partially Achieved | Not Achieved>'',
\newline
``Goal 2 (interaction\_outcome\_task)'': ``<Achieved | Partially Achieved | Not Achieved>''
\newline
\}
\newline
\}
\newline
\}

--------------------------------------------------

STRICT RULES:

- You MUST include ALL characters listed above.

- You must include ALL goals listed for each character.

- Do NOT merge multiple goals into one judgment.

- Do NOT add or remove characters.

- Do NOT include any text outside valid JSON.

- The output JSON MUST strictly follow the OUTPUT FORMAT schema.

- Do NOT include any additional fields (e.g., ``reasoning'', ``explanation'', ``steps'', or any other keys not defined in the schema).

- A goal can be marked as ``Achieved'' ONLY if ALL required steps and components defined in its evaluation rule are fully satisfied; otherwise it MUST be marked as ``Partially Achieved'' or ``Not Achieved''.

\end{promptbox}

\subsection{Int.-Out. Task Classification Prompt}
\label{app:Int.-Out.-task-classification-prompt}

This subsection presents the prompt we use to classify \textbf{Int.-Out.}.

\begin{promptbox}{Int.-Out. Classification Prompt}
\small
\ttfamily
\raggedright

System Prompt:

You are an expert evaluator of visual grounding in structured multi-agent social interaction tasks.

You are given:

- characters

- interaction\_outcome\_task for each character

==================== Task clarification ====================

You must analyze only the interaction\_outcome\_task field.

This is a direct two-class classification task.

For EACH character, classify the interaction\_outcome\_task into exactly one of the following two categories:

1) ``less vision-intensive task''

- Use this category when the outcome can be judged through dialogue, verbal acknowledgment, reasoning, claims, or interaction structure

- Use this category when the outcome contains no visual wording

- Use this category when the outcome only contains vague observable or behavioral wording

- Vague observable or behavioral wording includes terms such as:

- observable

- visible

- behavior

- shown through behavior

- hesitation

- withdrawal

- breakdown

- These terms do NOT count as concrete visual channels unless a specific visual channel is explicitly named

2) ``vision-intensive task''

- Use this category when the outcome explicitly depends on a concrete visual channel

- Concrete visual channels include:

- gaze / eye contact / gaze shifts / gaze avoidance

- facial expression (e.g., tightening, smiling, tension)

- visible emotion / affect (e.g., visible anxiety, tension)

- posture / body orientation (e.g., leaning, withdrawal, posture shifts)

==================== Decision rules ====================

- Only analyze interaction\_outcome\_task

- Be literal and conservative

- If a concrete visual channel appears, classify the task as ``vision-intensive task''

- If only vague observable, visible, or behavioral wording appears, classify the task as ``less vision-intensive task''

- If there is no visual, observable, or behavioral wording, classify the task as ``less vision-intensive task''

==================== Input format ====================

You will receive this JSON structure for one scene:

\{
\newline
``characters'': \{
\newline
``Character Name'': \{
\newline
``interaction\_outcome\_task'': ``...''
\newline
\}
\newline
\}
\newline
\}

==================== Output requirements ====================

For EACH character, return:

- category: exactly one of ``vision-intensive task'' or ``less vision-intensive task''

- short\_reason: a brief reason under 24 words

- evidence\_phrase: the exact phrase from interaction\_outcome\_task that supports the classification; use an empty string when the category is ``less vision-intensive task'' because no concrete visual phrase appears

Output strict JSON with no additional commentary.

==================== Return format ====================

\{
\newline
``characters'': \{
\newline
``Character Name'': \{
\newline
``category'': ``vision-intensive task | less vision-intensive task'',
\newline
``short\_reason'': ``<under 24 words>'',
\newline
``evidence\_phrase'': ``<exact phrase from interaction\_outcome\_task, or empty if no phrase is needed>''
\newline
\}
\newline
\}
\newline
\}

User Prompt:

Classify this scene:

\{\{ scene\_json \}\}

\end{promptbox}

\end{document}